\documentclass[10pt,twocolumn,letterpaper]{article}

\usepackage[pagenumbers]{cvpr} 

\definecolor{cvprblue}{rgb}{0.21,0.49,0.74}
\usepackage[pagebackref,breaklinks,colorlinks,allcolors=cvprblue]{hyperref}

\definecolor{headerblue}{HTML}{EAF2F8}
\definecolor{headergreen}{HTML}{E8F6F3}
\definecolor{headeryellow}{HTML}{FEF9E7}
\definecolor{headerpurple}{HTML}{F4ECF7}
\definecolor{linegray}{gray}{0.90}
\definecolor{mygreen}{HTML}{2E8B57}
\definecolor{myred}{HTML}{D22B2B}
\newcommand{\cmark}{\textcolor{mygreen}{\checkmark}}
\newcommand{\xmark}{\textcolor{myred}{\ding{55}}}

\usepackage{array}
\usepackage{amssymb}
\usepackage{pifont}         
\usepackage{booktabs}       
\usepackage{makecell}
\usepackage{multirow}
\usepackage[table, xcdraw]{xcolor}  
\usepackage{tabularx}

\def\paperID{} 
\def\confName{CVPR}
\def\confYear{2026}

\title{SegEarth-R2: Towards Comprehensive Language-guided Segmentation for Remote Sensing Images}

\author{
Zepeng Xin$^{1*}$, Kaiyu Li$^{1*}$, Luodi Chen$^{1}$, Wanchen Li$^{1}$, Yuchen Xiao$^{1}$, 
Hui Qiao$^{2}$,\\ Weizhan Zhang$^{1}$, Deyu Meng$^{1}$, Xiangyong Cao$^{1\dagger}$\\[4pt]
$^{1}$Xi'an Jiaotong University \quad $^{2}$ China Telecom Shaanxi Branch
}

\begin{document}
\twocolumn[{
\renewcommand\twocolumn[1][]{#1}
\maketitle
\vspace{-12mm}
    \begin{center}
    \centering
    \captionsetup{type=figure}
    \includegraphics[width=0.95\linewidth]{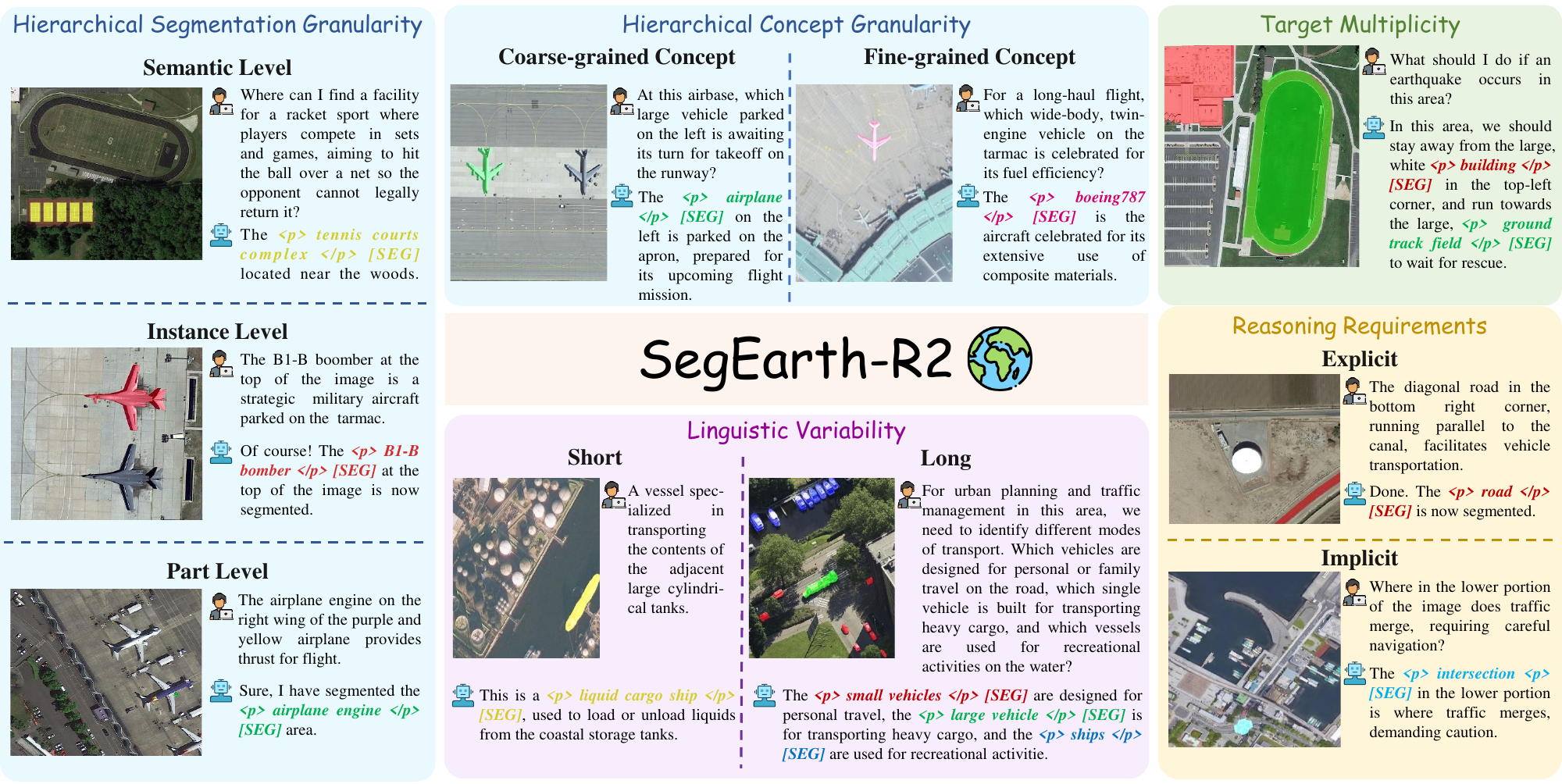}
    \captionof{figure}{SegEarth-R2 can handle various language-guided segmentation in remote sensing across four key dimensions: (1) \textbf{\textcolor[RGB]{45,83,150}{Hierarchical Granularity}}; (2) \textbf{\textcolor[RGB]{83,129,52}{Target Multiplicity}}; (3) \textbf{\textcolor[RGB]{187,137,0}{Reasoning Requirements}}; and (4) \textbf{\textcolor[RGB]{149,0,149}{Linguistic Variability}}.}
    \label{fig:examples}
\end{center}
}]

\renewcommand{\thefootnote}{*}
\footnotetext[1]{Equal contribution}
\renewcommand{\thefootnote}{\dag}
\footnotetext[2]{Corresponding author (caoxiangyong@mail.xjtu.edu.cn)}
\renewcommand{\thefootnote}{\arabic{footnote}}

\begin{abstract}
Effectively grounding complex language to pixels in remote sensing (RS) images is a critical challenge for applications like disaster response and environmental monitoring. Current models can parse simple, single-target commands but fail when presented with complex geospatial scenarios, e.g., segmenting objects at various granularities, executing multi-target instructions, and interpreting implicit user intent. To drive progress against these failures, we present LaSeRS, the first large-scale dataset built for comprehensive training and evaluation across four critical dimensions of language-guided segmentation: hierarchical granularity, target multiplicity, reasoning requirements, and linguistic variability. By capturing these dimensions, LaSeRS moves beyond simple commands, providing a benchmark for complex geospatial reasoning. This addresses a critical gap: existing datasets oversimplify, leading to sensitivity-prone real-world models. We also propose SegEarth-R2, an MLLM architecture designed for comprehensive language-guided segmentation in RS, which directly confronts these challenges. The model's effectiveness stems from two key improvements: (1) a spatial attention supervision mechanism specifically handles the localization of small objects and their components, and (2) a flexible and efficient segmentation query mechanism that handles both single-target and multi-target scenarios. Experimental results demonstrate that our SegEarth-R2 achieves outstanding performance on LaSeRS and other benchmarks, establishing a powerful baseline for the next generation of geospatial segmentation.
All data and code will be released at \href{https://github.com/earth-insights/SegEarth-R2}{\textcolor{cvprblue}{https://github.com/earth-insights/SegEarth-R2}}.
\end{abstract}


\section{Introduction}
\label{sec:intro}
Imagine you are walking in a park when the ground begins to shake violently—an earthquake. Immediately, instinct kicks in and your mind swiftly translates the abstract goal of ``finding a safe refuge'' into a complex, spatially-aware plan (\eg, the target multiplicity case depicted in Figure~\ref{fig:examples}): simultaneously identifying the nearby, open \textbf{\textit{\textcolor[RGB]{0,176,80}{ground track field}}} (the safe destination) while also urgently moving away from the surrounding \textbf{\textit{\textcolor[RGB]{192,0,0}{tall, swaying buildings}}} (the hazards). This remarkable human ability to ground a high-level, compound instruction into a precise, multi-target, pixel-level action is vital for survival. Yet, it is precisely this ability that even advanced AI systems for Earth observation struggle to replicate. While these systems excel at executing literal commands~\cite{liu2024rotated,dong2024cross,yuan2024rrsis,li2025segearth,shabbir2025geopixel,zhou2024geoground,ou2025geopix,lei2024exploring,li2025segearthov,li2025dynamicearth}, they fall short when interpreting abstract natural language that requires understanding spatial relationships between multiple targets, especially in the context of precise language-guided segmentation in complex, real-world remote sensing (RS) environments.

This inability to interpret complex instructions directly hinders the deployment of advanced geoinformation systems. Indeed, language-guided segmentation holds immense promise across diverse and complex real-world RS applications, including environmental dynamics, urban planning, and, critically, disaster management~\cite{rolf2024mission, lu2025vision, wang2025disasterm3, rahnemoonfar2021floodnet, danish2025geobenchvlm, chen2025rscc, an2024choice, li2025annotation}.
However, its development has been significantly inhibited by the lack of comprehensive training and evaluation resources. Specifically, existing datasets do not adequately capture the full range of required complexity, such as segmenting objects across hierarchical granularity and handling complex target multiplicity instructions.

To address these critical gaps, we introduce LaSeRS, a large-scale dataset of over 40k high-quality masks and 30k QA pairs, generated via a scalable semi-automated annotation pipeline.
As illustrated in Figure~\ref{fig:examples}, LaSeRS is constructed to serve as a comprehensive training and evaluation resource across the four dimensions of RS language-guided segmentation. Unlike prior works limited to single-target and explicit queries, LaSeRS systematically incorporates: \textbf{\textcolor[RGB]{45,83,150}{Hierarchical Granularity}}, requiring both conceptual (coarse- to fine-grained) and segmentation (semantic to part-level) precision; \textbf{\textcolor[RGB]{83,129,52}{Target Multiplicity}}, where a single instruction demands grounding of multiple objects; \textbf{\textcolor[RGB]{187,137,0}{Reasoning Requirements}}, both explicit descriptions and challenging implicit instructions; and \textbf{\textcolor[RGB]{149,0,149}{Linguistic Variability}}, spanning concise short queries and highly detailed long descriptions. Further, LaSeRS significantly pushes scale, offering more than 5$\times$ the category coverage of existing RS datasets~\cite{li2025segearth, liu2024rotated, dong2024cross, zhou2024geoground, livrsbench}, thereby significantly expanding the boundary of language-guided segmentation in RS.

The LaSeRS dataset poses some critical challenges to existing models~\cite{shabbir2025geopixel,li2025segearth, lai2024lisa, wei2024instructseg}, motivating the introduction of our SegEarth-R2 model. We built our model to overcome two primary failures. First, existing methods struggle with fine-grained segmentation of small and part-level object in LaSeRS. SegEarth-R2 confronts this with a spatial attention supervision mechanism, enforcing an internal focus on precise details. Second, grounding multiple targets from a single instruction is inefficient or unsupported by previous query designs~\cite{li2025segearth}. SegEarth-R2 address this with a flexible segmentation query mechanism capable of dynamically handling both single and multi-target requests. 



In summary, our contributions are as follows:
\begin{itemize}
    \item We introduce LaSeRS, the first benchmark for training and evaluation comprehensive language-guided segmentation in RS across four critical dimensions spanning hierarchical granularity, target multiplicity, reasoning requirements, and linguistic variability.
    \item We propose SegEarth-R2, a effective MLLM architecture for comprehensive language-guided segmentation in RS. It incorporates a spatial attention supervision for precise fine-grained localization and a flexible query mechanism for handling multi-target instructions.
   \item SegEarth-R2 sets a new state-of-the-art on LaSeRS and other public benchmarks. With only 3B parameters, it demonstrates superior performance against larger models, establishing a powerful yet highly efficient baseline for the comprehensive language-guided segmentation.
\end{itemize}


\begin{table*}[h]
    \centering
    \caption{Comparison between LaSeRS and other related datasets. TI: Textual Instruction, TA: Textual Answer, BBox: Bounding Box, Seg: Segmentation Mask. LaSeRS outperforms other datasets with the largest number of classes and unique support for \textbf{\textcolor[RGB]{45,83,150}{Hierarchical Granularity}}, \textbf{\textcolor[RGB]{83,129,52}{Target Multiplicity}}, \textbf{\textcolor[RGB]{187,137,0}{Reasoning Requirements}}, and \textbf{\textcolor[RGB]{149,0,149}{Linguistic Variability}}.}
    \vspace{-0.5em}
    \scalebox{0.68}
    {

\begin{tabular}{l c c c >{\centering\arraybackslash}p{1.5cm} >{\centering\arraybackslash}p{1.5cm} >{\centering\arraybackslash}p{1.5cm} >{\centering\arraybackslash}p{1.5cm} >{\centering\arraybackslash}p{1.5cm} >{\centering\arraybackslash}p{1.5cm} >{\centering\arraybackslash}p{1.5cm} >{\centering\arraybackslash}p{1.5cm}}
    \toprule
    \multirow{2}{*}{Dataset} & \multirow{2}{*}{Masks} & \multirow{2}{*}{Class Num} & \multirow{2}{*}{Annotation Type} & \multicolumn{2}{c}{\cellcolor{headerblue}Hierarchical Granularity} & \multicolumn{2}{c}{\cellcolor{headergreen}Target Multiplicity} & \multicolumn{2}{c}{\cellcolor{headeryellow}Reasoning Requirements} & \multicolumn{2}{c}{\cellcolor{headerpurple}Linguistic Variability} \\
    
    & & & & \multicolumn{1}{c}{\cellcolor{headerblue}Segmentation} & \multicolumn{1}{c}{\cellcolor{headerblue}Concept} & \multicolumn{1}{c}{\cellcolor{headergreen}Single} & \multicolumn{1}{c}{\cellcolor{headergreen}Multiple} & \multicolumn{1}{c}{\cellcolor{headeryellow}Explicit} & \multicolumn{1}{c}{\cellcolor{headeryellow}Implicit} & \multicolumn{1}{c}{\cellcolor{headerpurple}Long} & \multicolumn{1}{c}{\cellcolor{headerpurple}Short} \\
    \midrule

    EarthVQA~\cite{wang2024earthvqa} & 6,000 & 14 & TI, TA & \xmark & \xmark & \cmark & \cmark & \cmark & \cmark & \xmark & \cmark \\
    
    RegSegRS~\cite{yuan2024rrsis} & 4,420 & 14 & TI, Seg & \xmark & \xmark & \cmark & \xmark & \cmark & \xmark & \xmark & \cmark \\
    
    RRSIS-D~\cite{liu2024rotated} & 17,420 & 20 & TI, Seg, BBox & \xmark & \xmark & \cmark & \xmark & \cmark & \xmark & \xmark & \cmark \\

    RISBench~\cite{dong2024cross} & 52,472 & 26 & TI, Seg, BBox & \xmark & \xmark & \cmark & \xmark & \cmark & \xmark & \xmark & \cmark \\

    EarthReason~\cite{li2025segearth} & 5,434 & 28 & TI, TA, Seg & \xmark & \xmark & \cmark & \xmark & \xmark & \cmark & \cmark & \xmark \\ 

    \midrule
    
    LaSeRS & 40,396 & 122 & TI, TA, Seg, BBox & \cmark & \cmark & \cmark & \cmark & \cmark & \cmark & \cmark & \cmark \\
    \bottomrule
\end{tabular}}
    \label{tab:dataset_compare}
\end{table*}

\section{Related works}
\label{sec:formatting}
\noindent \textbf{Multimodal Large Language Models (MLLMs).} MLLMs build upon Large Language Models (LLMs) to acquire vision capabilities. In the general domain, pioneers like LLaVA~\cite{liu2023visual, li2024llava, liu2024improved} and InstructBLIP~\cite{dai2023instructblip} aligned visual features with language representations using a vision-language connector, enhanced by instruction tuning. While these initial models focused on image-level understanding, subsequent works like GPT4RoI~\cite{zhang2024gpt4roi} and RegionGPT~\cite{guo2024regiongpt} advanced this by introducing regional understanding.
This evolution is mirrored in the RS domain, where MLLMs have evolved from foundational conversational agents like RSGPT~\cite{hu2025rsgpt}. Subsequent RS-specific models have integrated more complex functionalities. For instance, GeoChat~\cite{kuckreja2024geochat} supports region-specific inputs and visual grounding through oriented bounding box coordinates in its responses. Others extend capabilities to RS-video captioning, such as SkyEyeGPT~\cite{zhan2025skyeyegpt}, or integrate multi-sensor tasks, as exemplified by EarthGPT~\cite{zhang2024earthgpt} and EarthDial~\cite{soni2025earthdial}. However, despite these advances in interpretation~\cite{pang2024h2rsvlm, bazi2024rs, irvin2024teochat, li2025lhrs} and relation reasoning~\cite{zhan2025skyeyegpt, wang2025geollava, pang2025vhm}, a fundamental limitation persists: these models lack crucial pixel-level understanding and segmentation capabilities.

\noindent \textbf{Language-guided Segmentation.} Language-guided segmentation refers to locating target regions at the pixel level from natural language, including referring segmentation~\cite{ding2021vision,liu2023polyformer,wang2022cris,yang2022lavt, zong2025ground, hu2025groundingsuite, xiang2025advancing} and reasoning segmentation~\cite{lai2024lisa,ren2024pixellm, jang2025mmr, zong2025ground}. In RS, RefSegRS~\cite{yuan2024rrsis} introduced the first referring segmentation dataset, while RRSIS-D~\cite{liu2024rotated} and some subsequent works~\cite{dong2024cross,yang2025large, zhou2024geoground,ou2025geopix, zhou2024geoground, ou2025geopix} expanded dataset scale and enriched prompts, though still limited to explicit category queries. EarthReason~\cite{li2025segearth} further introduced geospatial pixel reasoning with the first reasoning segmentation dataset for Earth observation. Yet, while a valuable step, it primarily focuses on implicit reasoning for single targets, leaving the complex challenges of multi-granularity (\eg, part-level and fine-grained concept) and multi-target grounding unaddressed. To bridge this gap, LaSeRS offers a more comprehensive benchmark that systematically incorporates all of these critical dimensions.

\noindent \textbf{MLLMs for language-guided segmentation.} Recent work has explored leveraging the reasoning capabilities of MLLMs for complex language-guided segmentation. In the general domain, a dominant approach employs MLLMs as a reasoning engine to provide prompt embeddings (\eg, [SEG] tokens) for SAM~\cite{kirillov2023segment, ravi2024sam}, thereby integrating reasoning into mask generation~\cite{lai2024lisa, jang2025mmr, xia2024gsva, yan2024visa, wei2024lasagna, yuan2025sa2va, han2025groundingface}. Other methods follow pipelines inspired by Mask2Former~\cite{cheng2022masked} (``propose-then-select")~\cite{zhang2024psalm, wei2024instructseg, wei2024hyperseg, wang2025x, xiang2025advancing} or treat segmentation as text generation~\cite{lan2024text4seg, lan2025text4seg++}. However, these general-purpose methods often underperform on RS datasets, which are characterized by the prevalence of small objects and significant scale variations. This has spurred the development of specialized MLLMs for the RS field~\cite{liu2024rsunivlm, zhou2024geoground, ou2025geopix, ma2025geomag, yao2025falcon, shu2025earthmind}, with methods like GeoPixel~\cite{shabbir2025geopixel} adapting grounded conversation, SegEarth-R1~\cite{li2025segearth} and LISAt~\cite{quenum2025lisat} handling implicit queries, Geo-R1~\cite{zhang2025geo} and RemoteReasoner~\cite{yao2025remotereasoner} exploring reinforcement learning. Despite these advances in RS-specific models, they still lack the ability to handle language-guided segmentation in complex scenarios, such as multi-target settings or multi-granularity (\eg, part-level).
\begin{figure*}[t]
  \centering
\includegraphics[width=1.0\textwidth]
{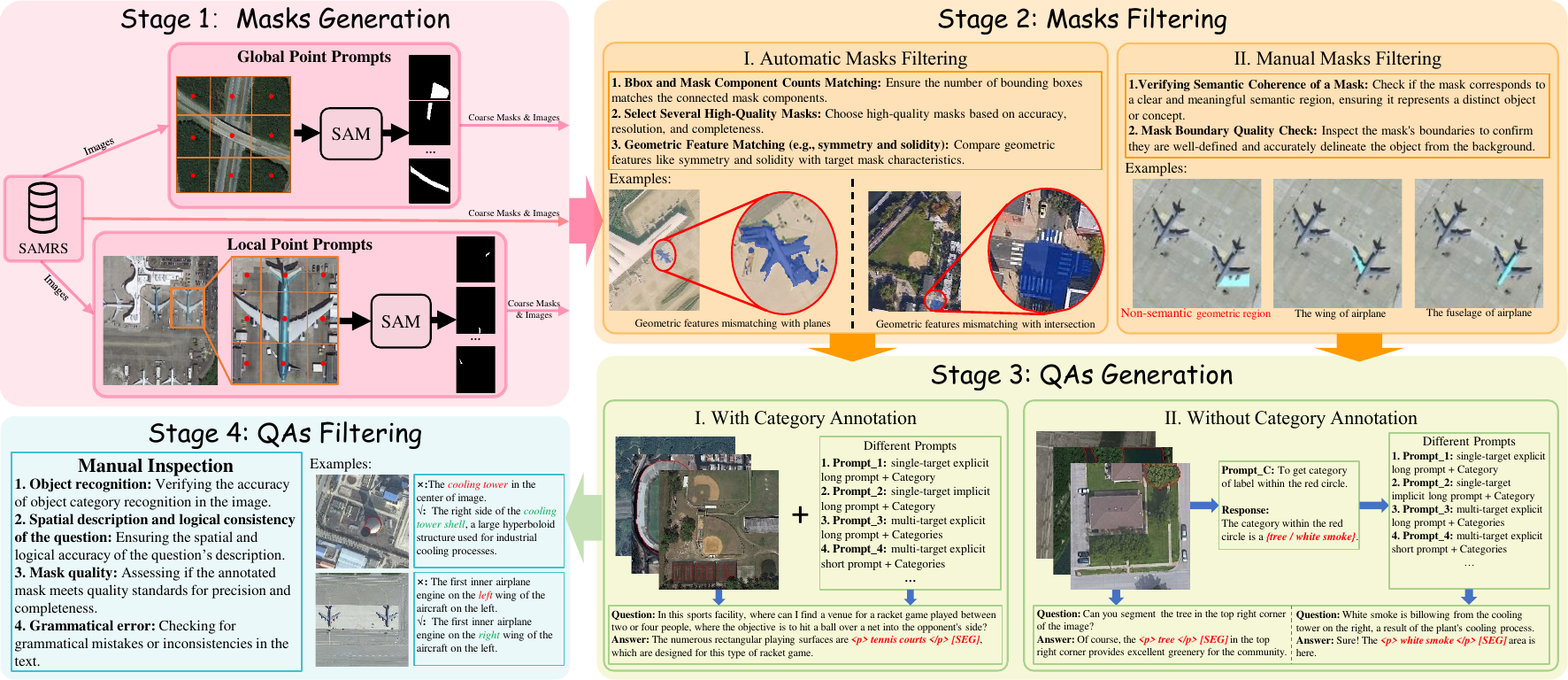}
\vspace{-1.5em}
  \caption{Overview of the construction of LaSeRS. Our scalable, semi-automatic annotation pipeline comprises four stages: two dedicated to data generation and two dedicated to filtering. Data quality is rigorously ensured through both automatic and manual verification.}
  \label{fig:pipeline}
\end{figure*}

\begin{figure}[t]
    \centering
    \includegraphics[width=0.45\textwidth]{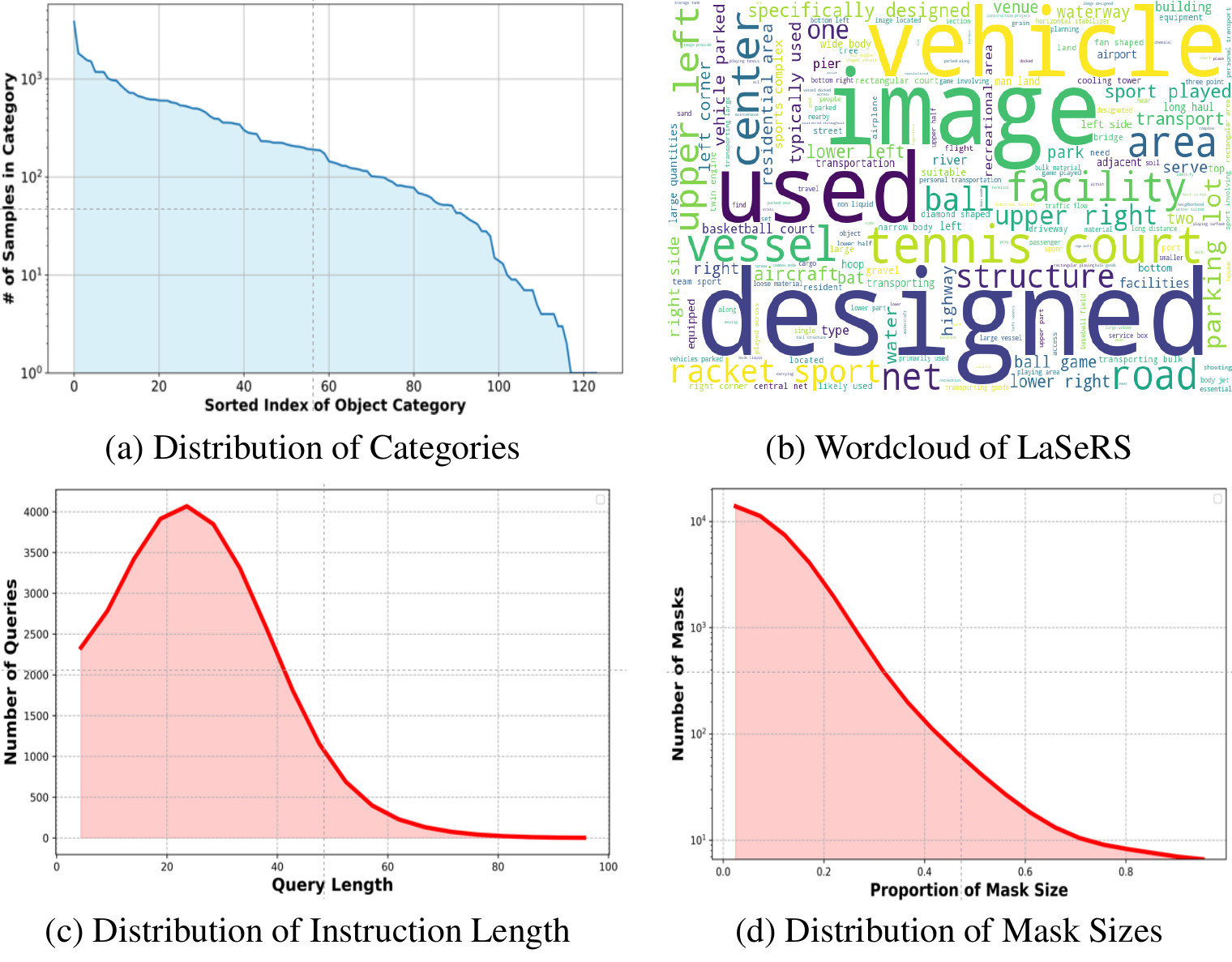}
    \vspace{-0.5em}
    \caption{
    Data Statistics of LaSeRS.
    }
    \label{fig:statistics}
    \vspace{-4mm}
\end{figure}

\section{LaSeRS Dataset} 
\subsection{Data Generation and Filtering}
We develop a four-stage semi-automatic data generation pipeline to generate question-answer-mask triples, as shown in Figure~\ref{fig:pipeline}. More details about the dataset construction can be found in the Suppl.~\ref{sec:more_information}.

\noindent \textbf{Stage 1: Masks Generation.} Our mask annotations are sourced from two complementary workflows. First, we incorporate masks from the existing SAMRS~\cite{SAMRS} dataset. However, these annotations exhibit varying quality and are limited to $\sim$60 categories. Second, to expand coverage and introduce part-level details, we use SAM~\cite{kirillov2023segment} to generate masks for the entire image using a grid of point prompts. Then, for common RS object categories like airplanes and ships, we cropped the object region and apply a denser grid of prompts within that region to produce more refined, part-level masks. This stage generates a large volume of coarse masks that require subsequent filtering.

\noindent \textbf{Stage 2: Masks Filtering.} To ensure annotation quality, we apply distinct filtering methods based on the mask source. For masks originating from SAMRS, we discard samples where the number of connected components in a mask does not match the number of provided bounding box annotations. Given that objects in remote sensing images often exhibit clear geometric properties (\eg, the pronounced symmetry of airplanes and the rotational invariance typical of intersections), we then conduct geometric feature matching to filter out candidates that deviate significantly from a curated set of ideal, category-specific reference shapes (details in the Suppl.~\ref{sec:mask_filtering}). In contrast, all masks generated by the second workflow are manually reviewed. Annotators assess each mask based on two criteria: semantic coherence, to ensure it represents a meaningful object, and high boundary quality, to ensure mask precision. This stage yields approximately 50K high-quality, semantically meaningful masks at varying segmentation granularities, from semantic- and instance-level to part-level.

\noindent \textbf{Stage 3: QAs Generation.} We used Gemini-2.5-pro~\cite{comanici2025gemini} to create question-answer pairs, with the process differing based on the mask source. For masks with pre-existing category annotations from SAMRS, we leverage this information to guide the generation of diverse textual annotations. Conversely, for unlabeled masks from our second workflow, we first employ a prompting strategy to assign a category to each mask. This assigned category subsequently guides the generation of varied textual annotations. In both cases, we utilize multiple prompt templates to create question-answer pairs of varying styles, including explicit vs. implicit and detailed vs. concise. This stage ultimately yields approximately 40K question-answer pairs. However, due to potential hallucinations from the Gemini-2.5-pro, these pairs require further filtering.

\noindent \textbf{Stage 4: QAs Filtering.} Finally, all generated question-answer pairs are manually reviewed. This review verifies the accuracy of object recognition, the logical consistency of spatial descriptions, the quality of the associated masks, and checks for grammatical errors to ensure the reliability of the final dataset. This final filtering stage yields 30,830 high-quality question-answer-masks triples. Additionally, to accommodate the demand for coarse localization, we convert the masks into bounding box annotations via mask-to-bbox~\cite{liu2024remoteclip}.

\subsection{Data Statistics}
As detailed in Table~\ref{tab:dataset_compare}, LaSeRS is a large-scale dataset comprising 40,396 high-quality masks. The dataset's annotations span four types: textual instructions, textual answers, segmentation masks, and bounding boxes for coarse localization. The test set consists of 1,900 held-out masks and is meticulously partitioned to evaluate performance across four critical dimensions: \textbf{\textcolor[RGB]{45,83,150}{Hierarchical Segmentation Granularity}} (semantic, instance, and part-level), \textbf{\textcolor[RGB]{83,129,52}{Target Multiplicity}} (single, multiple), \textbf{\textcolor[RGB]{187,137,0}{Reasoning Requirements}} (explicit, implicit), and \textbf{\textcolor[RGB]{149,0,149}{Linguistic Variability}} (short, long).
Figure~\ref{fig:statistics} summarizes the statistical properties of LaSeRS. Panel (a) shows the distribution across 122 object categories, highlighting our dataset's broad coverage. This ensures the QA pairs are not skewed towards specific categories and exhibit high diversity. Further details on the categories are provided in the Suppl.~\ref{sec:category}. Panel (b) presents a word cloud of the instruction vocabulary. Panel (c) illustrates the instruction length distribution, ranging from just a few words to over 80, ensuring linguistic variability. Finally, panel (d) highlights the distribution of mask sizes, with a significant scale range from small, fine-grained parts to large regions exceeding 80\% of the image area. 

\begin{figure*}[t]
  \centering
\includegraphics[width=1.0\textwidth]
{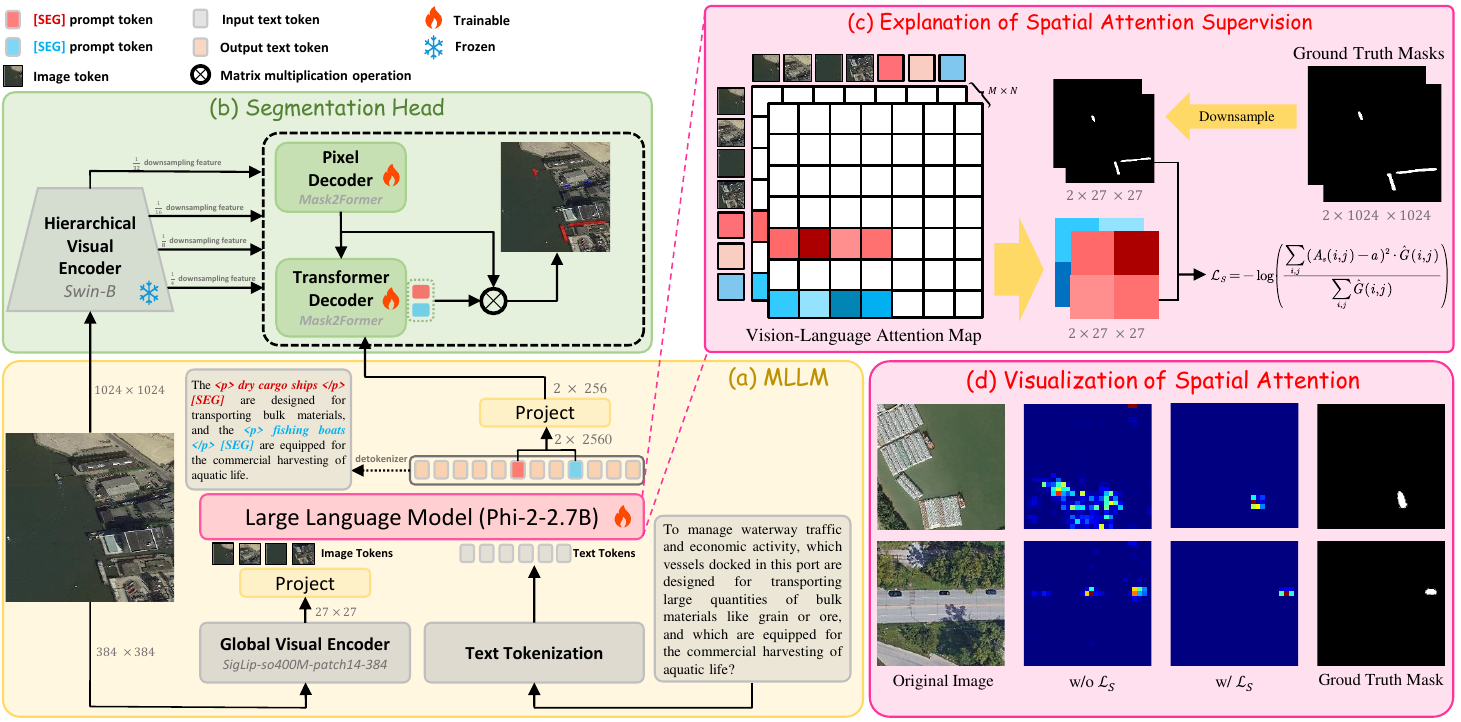}
  \caption{The overall architecture of SegEarth-R2. (a) is an MLLM for interpreting the image and instruction, (b) is the segmentation head, which includes a hierarchical visual encoder for generating multi-scale image features, and two decoders for producing masks (mask decoders). (c) illustrates the formation process of the spatial attention supervision. (d) presents a visualization of the attention map from the 6th head of the 16th layer in the MLLM, with additional visual results in the Suppl.~\ref{sec:spatial_attention}}
  \label{fig:overview}
\end{figure*}

\section{SegEarth-R2}
To address the challenges of language-guided segmentation in RS, we introduce SegEarth-R2, a framework designed for precise, language-driven segmentation in RS images. As depicted in Figure~\ref{fig:overview}, the model is composed of two primary components. The first is an MLLM that processes the input image and textual instruction to perform reasoning and generate a textual response. The second is a dedicated segmentation head that interprets the MLLM's internal semantic representation, specifically the [SEG] token queries, to produce pixel-level masks. The performance of SegEarth-R2 relies on two key mechanisms corresponding to the above components: a spatial attention supervision mechanism applied to the model's internal attention and a flexible and efficient segmentation query mechanism for mask generation, which are detailed in the following.
\subsection{Spatial Attention Supervision}
A defining characteristic of RS images is the extreme variation in object scales, where targets can range from vast geographical features to entities spanning only a few pixels (\eg, the small engine of an airplane). This presents a significant challenge for standard training paradigms. Supervision that relies solely on the final segmentation mask proves insufficient for instilling precise spatial awareness, as the learning signal becomes severely diluted when backpropagated to the network's shallower layers. This dilution effect often leads to poor localization of small or complexly shaped objects, as visualized in Figure~\ref{fig:overview}(d).

To counteract this, we introduce a spatial attention supervision mechanism, which provides direct supervisory signals to the model's internal vision-language representations. Rather than inferring focus from the final output, our approach constitutes a direct intervention within the model's reasoning pathway, compelling it to distinguish between foreground and background regions at the attention level. This process is illustrated in Figure~\ref{fig:overview}(c).

This method operates on the attention maps originating from the [SEG] token to the image patch tokens within the MLLM's transformer layers. Let $A_{(m,n)} \in \mathbb{R}^{d \times d}$ denote the attention map for the $n$-th head in the $m$-th transformer block, where the model has a total of $M$ blocks and $N$ heads per block, $d \times d$ represents the spatial dimensions of the image feature grid to which attention is applied. We then aggregate these individual maps into a unified attention grid by averaging them across all layers and heads, \ie, $A_S = \frac{1}{MN} \sum_{m=1}^{M} \sum_{n=1}^{N} A_{(m,n)}$. Using a downsampled version of the ground truth mask, $\hat{G} \in \{0,1\}^{d \times d}$, we calculate the average attention score, $a$, across all background regions:

\begin{equation}
a = \frac{\sum_{i,j} A_S(i,j) \cdot (1 - \hat{G}(i,j))}{\sum_{i,j} (1 - \hat{G}(i,j))}
\nonumber
\end{equation}
where $A_S(i,j)$ and $\hat{G}(i,j)$ represent the $i$-th row and $j$-th column of the corresponding tensors.

The spatial attention supervision loss, $\mathcal{L}_S$, is then formulated to enforce a maximal separation between the attention scores on the foreground regions and the calculated background average $a$:
\begin{equation}
\mathcal{L}_S = -\log \left( \frac{\sum_{i,j} (A_{S}(i,j) - a)^2 \cdot \hat{G}(i,j)}{\sum_{i,j} \hat{G}(i,j)} \right)
\nonumber
\end{equation}

By maximizing the dissimilarity, this loss directly encourages the model to sharpen its focus specifically on the target area. This provides a clear and localized learning signal to the intermediate layers, substantially improving the model's overall ability to ground and segment challenging objects with remarkable high fidelity.

\begin{table*}[t]
    \centering
    \caption{Performance comparison on LaSeRS. Baselines including LISA, PixelLM, $M^2$A, and GeoPixel are fine-tuned directly on LaSeRS. GLaMM-ft is pre-trained on large-scale natural image segmentation datasets before fine-tuning. Metrics are gIoU/cIoU. \textbf{Bold} and \underline{underlined} indicate the best and second-best scores, respectively.}
    \scalebox{0.76}
    {

\begin{tabular}{l|| >{\centering\arraybackslash}p{1.52cm} >{\centering\arraybackslash}p{1.52cm} >{\centering\arraybackslash}p{1.52cm} |>{\centering\arraybackslash}p{1.52cm} >{\centering\arraybackslash}p{1.52cm}|>{\centering\arraybackslash}p{1.52cm} >{\centering\arraybackslash}p{1.52cm}|>{\centering\arraybackslash}p{1.52cm} >{\centering\arraybackslash}p{1.52cm}|c}
    \toprule
    \multirow{2}{*}{Model} & \multicolumn{3}{c|}{\cellcolor{headerblue}{Hierarchical Segmentation Granularity}} & \multicolumn{2}{c|}{\cellcolor{headergreen}{Target Multiplicity}} & \multicolumn{2}{c|}{\cellcolor{headeryellow}{Reasoning Requirements}} & \multicolumn{2}{c|}{\cellcolor{headerpurple}{Linguistic Variability}} & \multirow{2}{*}{Avg.} \\
     & \cellcolor{headerblue}{Semantic} & \cellcolor{headerblue}{Instance} & \cellcolor{headerblue}{Part} & \cellcolor{headergreen}{Single} & \cellcolor{headergreen}{Multiple} & \cellcolor{headeryellow}{Explicit} & \cellcolor{headeryellow}{Implicit} & \cellcolor{headerpurple}{Short} & \cellcolor{headerpurple}{Long} & \\
    \midrule
    LISA-7B~\cite{lai2024lisa} & 26.4/23.2 & 20.5/25.0 & 16.1/11.6 & 37.3/32.2 & 18.2/22.4 & 27.1/24.3 & 21.5/25.6 & 34.1/27.8 & 38.4/33.9 & 26.6/25.1 \\
    LISA-13B~\cite{lai2024lisa} & 27.0/24.5 & 22.3/25.6 & 17.7/13.1 & 38.4/34.2 & 19.9/23.5 & 27.1/25.5 & 22.6/25.8 & 35.2/28.0 & 38.4/34.3 & 27.6/26.1 \\
    PixelLM-7B~\cite{ren2024pixellm} & 32.0/32.8 & 26.6/30.0 & 13.2/16.5 & 44.3/40.4 & 20.2/23.5 & 25.0/23.1 & 23.9/21.9 & 41.6/38.9 & 37.1/34.5 & 29.3/29.1 \\
    PixelLM-13B~\cite{ren2024pixellm} & 31.6/34.0 & 27.5/30.2 & 15.8/17.6 & 42.2/40.5 & 20.9/22.4 & 26.3/24.4 & 25.9/22.1 & 42.0/39.1 & 37.1/34.5 & 29.9/29.4 \\		
    GLaMM-ft-7B~\cite{hanoona2023GLaMM} & 44.8/47.9 & 41.2/48.3 & 32.6/42.7 & 47.3/\underline{50.3} & 32.2/41.0 & 59.1/60.3 & \underline{42.6}/44.8 & 50.4/54.8 & 42.6/44.8 & 43.6/48.3 \\
    $M^2$A-7B~\cite{jang2025mmr} & 30.1/33.0 & 23.0/24.8 & 18.6/17.2 & 45.4/37.6 & 20.9/24.8 & 35.8/30.4 & 23.3/26.7 & 35.8/32.8 & 41.5/36.7 & 30.5/29.3 \\
    GeoPixel-8B~\cite{shabbir2025geopixel} & \underline{51.4}/\underline{57.2} & \underline{44.1}/\underline{49.3} & \underline{43.9}/\underline{52.4} & \underline{55.0}/45.8 & \textbf{49.2}/\underline{49.7} & \underline{66.5}/\underline{61.3} & 41.1/\underline{58.3} & \underline{51.1}/\underline{59.3} & \textbf{51.4}/\underline{63.2} & \underline{50.4}/\underline{55.2} \\
    \midrule
    \cellcolor{linegray}{SegEarth-R2-3B} & \cellcolor{linegray}{\textbf{60.2}/\textbf{71.8}} & \cellcolor{linegray}{\textbf{65.4}/\textbf{70.3}} & \cellcolor{linegray}{\textbf{64.8}/\textbf{68.3}} & \cellcolor{linegray}{\textbf{55.1}/\textbf{69.2}} & \cellcolor{linegray}{\underline{38.3}/\textbf{56.2}} & \cellcolor{linegray}{\textbf{78.4}/\textbf{80.4}} & \cellcolor{linegray}{\textbf{42.8}/\textbf{59.7}} & \cellcolor{linegray}{\textbf{60.2}/\textbf{69.9}} & \cellcolor{linegray}{\underline{50.1}/\textbf{65.7}} & \cellcolor{linegray}{\textbf{57.2}/\textbf{67.9}} \\
    \bottomrule
\end{tabular}}
    \label{tab:main_results}
\end{table*}

\subsection{Segmentation Query Mechanism}
Previous approaches, such as InstructSeg~\cite{wei2024instructseg}, rely on a cumbersome ``propose-then-select" strategy. They first generate a large pool of candidate masks (\eg, 100) and are subsequently plagued by a redundant and time-consuming matching process to filter for the desired output; this design is not only computationally inefficient but also inflexible. On the other hand, while SegEarth-R1's~\cite{li2025segearth} ``instruction-as-query" method offers simplicity, it is inherently limited by its assumption of a single target per instruction. This naive one-to-one mapping makes it fundamentally unsuitable for complex, multi-target scenarios.

As illustrated in Figure~\ref{fig:overview}, to handle multi-target scenarios, we adopt the approach from LISA~\cite{lai2024lisa} by introducing a special [SEG] token. The model learns to dynamically output multiple [SEG] tokens based on the instruction's context (\eg. ``\textit{In this area, we should stay away from the large, white \textless p\textgreater building \textless /p\textgreater [SEG] in the top-left corner, and run towards the large, \textless p\textgreater ground track field \textless /p\textgreater [SEG] to wait for rescue.}"), which then function as individual segmentation queries. Our architecture is composed of a Hierarchical Visual Encoder, \ie, the Swin-B~\cite{liu2021swin} which extracts multi-level image features. These features are then processed by a Pixel Decoder and a Transformer Decoder; we note that both of these modules adopt the identical architecture from Mask2Former~\cite{cheng2022masked}. The Pixel Decoder fuses the hierarchical features from the Swin-B, and the Transformer Decoder utilizes the [SEG] token queries to interact with these fused features, generating the final masks. This flexible and efficient segmentation query mechanism not only evades the redundant matching of ``propose-then-select" methods but also capably handles the complex multi-target scenarios where ``instruction-as-query" methods fail.
\section{Experiment}
\subsection{Implementation Details}
\textbf{Training objectives.} We train SegEarth-R2 with a unified loss function $\mathcal{L}$, which is the weighted sum of four distinct components:
\begin{gather*}
    \mathcal{L} = \mathcal{L}_t + \mathcal{L}_b + \mathcal{L}_d + \lambda_{S} \mathcal{L}_{S}
\end{gather*}

Here, $\mathcal{L}_{t}$ is the standard autoregressive cross-entropy loss for text generation. For mask supervision, we combine the per-pixel binary cross-entropy loss $\mathcal{L}_{b}$ and DICE loss $\mathcal{L}_{d}$, while $\mathcal{L}_{S}$ denotes the spatial attention supervision, which provides direct supervision on the model's internal attention and is controlled by the coefficient $\lambda_{S}$

\noindent\textbf{Experiment Settings.} We use the pre-trained lightweight MLLM Mipha-3B~\cite{zhu2024mipha} as our base model. Following SegEarth-R1~\cite{li2025segearth}, we use Swin-B~\cite{liu2021swin} as the hierarchical visual encoder and the mask decoder from Mask2Former~\cite{cheng2022masked}. Furthermore, to conserve GPU memory, we keep all Visual Encoders frozen and employ LoRA with a rank of 8 to fine-tune the LLM, while the smaller Pixel Decoder and Transformer Decoder components are fully fine-tuned. More training details can be found in Suppl.~\ref{sec:training_details}.

\noindent\textbf{Metrics.} We use gIoU and cIoU as evaluation metrics, following prior studies~\cite{wu2020phrasecut,lai2024lisa}.

\begin{table}[t]
    \centering
    \caption{Performence comparison on three RS referring segmentation benchmarks. To ensure a fair comparison, our results are obtained by training separately on each of the three training set. The metric is gIoU.}
    \scalebox{0.70}{
\begin{tabular}{ll||cc|cc|cc}
    \toprule
    \multirow{2}{*}{Method} & \multirow{2}{*}{Pub} & \multicolumn{2}{c|}{RRSIS-D} & \multicolumn{2}{c|}{RefSegRS} & \multicolumn{2}{c}{RISBench} \\
    & & val & test & val & test & val & test \\
    \midrule
    \multicolumn{8}{l}{\textit{Segmentation Specialists}} \\
    \addlinespace
    CMPC+~\cite{liu2021cross} & TPAMI'21 & 51.4 & 50.2 & 47.1 & 43.7 & 45.8 & 46.7 \\
    RIS-DMMI~\cite{hu2023beyond} & CVPR'23 & 60.7 & 60.1 & 65.7 & 52.2 & \underline{62.6} & \underline{63.9} \\
    LAVT~\cite{yang2022lavt} & CVPR'22 & 61.5 & 61.0 & 61.5 & 47.4 & 60.5 & 61.9 \\
    RMSIN~\cite{liu2024rotated} & CVPR'24 & 65.1 & 64.2 & 73.8 & 62.6 & 61.8 & 63.1 \\
    \midrule
    \multicolumn{8}{l}{\textit{MLLM based segmentation}} \\
    \addlinespace 
    LISA~\cite{lai2024lisa} & CVPR'24 & 27.8 & 26.8 & - & - & - & - \\
    PixelLM~\cite{ren2024pixellm} & CVPR'24 & 33.9 & 31.7 & - & - & - & - \\
    GeoGround~\cite{zhou2024geoground} & arXiv'24 & 61.1 & 60.5 & - & - & - & - \\
    GeoPixel~\cite{shabbir2025geopixel} & ICML'25 & \underline{68.0} & \underline{67.3} & - & - & - & - \\
    SegEarth-R1~\cite{li2025segearth} & arXiv'25 & 67.6 & 66.4 & \underline{82.2} & \underline{72.5} & - & - \\
    Text4Seg++~\cite{lan2025text4segadvancingimagesegmentation} & arXiv'25 & 64.1 & 62.8 & - & - & - & - \\
    \rowcolor{linegray}
    SegEarth-R2 &  & \textbf{68.8} & \textbf{67.9} &  \textbf{84.4} & \textbf{74.8}  & \textbf{69.8} & \textbf{70.5}  \\ 
    \bottomrule
\end{tabular}}
    \label{tab:otherbenchmark}
\end{table}

\begin{table}[t]
    \centering
    \caption{Performence comparison on RS reasoning segmentation benchmark: EarthReason~\cite{li2025segearth}. Our model is trained solely on the EarthReason training set.}
    \scalebox{0.70}{%
\begin{tabular}{l c ccccc}
    \toprule
    \multirow{2}*{Method} & \multirow{2}*{LLM} & \multicolumn{2}{c}{Val} & \multicolumn{2}{c}{Test} & \multirow{2}*{Avg.} \\
    & & gIoU & cIoU & gIoU & cIoU & \\
    \midrule
    LISA~\cite{lai2024lisa} & Vicuna-7B & 61.0 & 57.4 & 60.9 & 59.1 & 59.6 \\
    PixelLM~\cite{ren2024pixellm} & Vicuna-7B & 57.9 & 57.8 & 60.0 & 59.2 & 58.7 \\
    SegEarth-R1~\cite{li2025segearth} & Phi-1.5-1.3B & 68.6 & 64.1 & 70.8 & 68.3 & 68.0 \\
    RemoteReasoner~\cite{yao2025remotereasoner} & Qwen2.5-7B & 69.0 & 67.8 & 71.0 & \underline{69.1} & 69.2 \\
    Text4Seg++~\cite{lan2025text4segadvancingimagesegmentation} & Qwen2-7B & \underline{71.9} & \textbf{69.8} & \underline{73.0} & 65.6 & \underline{70.1} \\
    \midrule
    \cellcolor{linegray}{SegEarth-R2-3B} & \cellcolor{linegray}{Phi-2-2.7B} & \cellcolor{linegray}{\textbf{72.3}} & \cellcolor{linegray}{\underline{68.1}} & \cellcolor{linegray}{\textbf{73.5}} & \cellcolor{linegray}{\textbf{69.5}} & \cellcolor{linegray}{\textbf{70.9}} \\
    \bottomrule
\end{tabular}}
    \label{tab:earthreason_v1}
\end{table}

\subsection{Main Results}
As shown in Table~\ref{tab:main_results}, we conduct a comprehensive and fair evaluation on the LaSeRS benchmark, analyzing performance across four key dimensions.

\noindent \textbf{\textcolor[RGB]{45,83,150}{Hierarchical Segmentation Granularity.}} We observe a consistent performance degradation across nearly all models as the segmentation granularity refines, \ie, from the semantic-level, to the instance-level, and finally to the part-level, which often spans only a few pixels. This trend is expected, as finer granularities necessitate more precise model localization. Benefiting from our proposed spatial attention supervision mechanism, our model achieves the best performance across all granularities. This superiority is particularly pronounced at the challenging part-level, where our model surpasses the second-best method by a significant 20-point margin.

\noindent \textbf{\textcolor[RGB]{83,129,52}{Target Multiplicity.}} For complex multi-target scenarios, we observe a significant performance degradation across most models. GeoPixel is a notable exception, which shows much less degradation, a result we attribute to the inherent strength of its 8B MLLM foundation~\cite{zhang2024internlm}. We hypothesize that our model's 3B size limits its performance in these complex multi-target scenarios. Even so, our model still secures the best performance on single-target tasks and the second-best on multi-target tasks, significantly outperforming other 7B and even 13B models. 

\noindent \textbf{\textcolor[RGB]{187,137,0}{Reasoning Requirements.}} All models exhibit significantly better performance on explicit instructions compared to implicit ones. This is an expected outcome, as implicit instructions, unlike their explicit counterparts, require the model to infer the target segmentation region by reasoning over the instruction's context and its internal geographic knowledge. Notably, our model achieves excellent performance on both explicit and implicit instructions. 

\noindent \textbf{\textcolor[RGB]{149,0,149}{Linguistic Variability.}} Regarding linguistic variability, we observe an interesting phenomenon: models respond differently to instruction length. Specifically, LISA, $M^2$A, and GeoPixel excel at processing long, detailed instructions. In contrast, PixelLM, GLaMM, and SegEarth-R2 show more prominent performance on short, concise instructions. We hypothesize this divergence stems from fundamental differences in their model architectures. Notably, SegEarth-R2 demonstrates strong robustness, achieving superior performance on both instruction types.

\noindent \textbf{Overall. } With only 3B parameters, SegEarth-R2 significantly outperforms all other models on average, a result that showcases the superiority of our architecture. More qualitative results can be found in the Suppl.~\ref{sec:LaSeRS}. 

\subsection{Results on Existing Benchmarks}
To ensure SegEarth-R2's effectiveness is not limited to our proposed benchmark, we assess its generalization capabilities on established public datasets. For fair comparison, the model was trained separately on each benchmark's official training split.

\noindent \textbf{RS Referring Segmentation. } We evaluate on three referring segmentation benchmarks: RRSIS-D~\cite{liu2024rotated}, RefSegRS~\cite{yuan2024rrsis}, and RISBench~\cite{dong2024cross}. As shown in Table~\ref{tab:otherbenchmark}, SegEarth-R2 demonstrates strong cross-dataset generalization. On the RRSIS-D dataset, it outperforms the powerful 8B model, GeoPixel. On RefSegRS, SegEarth-R2 surpasses the previous state-of-the-art method by over 2 points. Furthermore, our model achieves the best performance on the larger-scale RISBench dataset. Qualitative results are provided in the Suppl.~\ref{sec:referring}.

\noindent \textbf{RS Reasoning Segmentation. } On the more challenging EarthReason~\cite{li2025segearth} benchmark for implicit reasoning segmentation in RS, as shown in Table~\ref{tab:earthreason_v1}, our model achieves an average score of 70.9, outperforming the previous state-of-the-art, Text4Seg++ (70.1), which represents masks using text. It also surpasses RemoteReasoner, a method exploring reinforcement learning for RS reasoning segmentation, demonstrating the superiority of our architecture in processing complex implicit instructions. Qualitative results are provided in Suppl.~\ref{sec:earthreason}.

\subsection{Ablation Study}
We conduct thorough ablation studies on our proposed spatial attention supervision ($\mathcal{L}_S$) and the adopted segmentation query mechanism. Furthermore, we investigate different combinations of the Segmentation Head (shown in Figure~\ref{fig:overview}) to demonstrate the superiority of our architecture.

\noindent \textbf{Spatial Attention Supervision Effect.} The coefficient $\lambda_{S}$ controls the weight of spatial attention supervision $\mathcal{L}_S$. We perform an ablation study on $\lambda_{S}$, and the results are shown in Table~\ref{tab:ablation_1}. As shown, setting $\lambda_{S}=0$, which removes this supervision, results in the worst performance, highlighting the necessity of spatial attention guidance. However, we observe that excessively large values of $\lambda_{S}$ (\eg, 0.1 or 0.05) degrade performance. We hypothesize this occurs because overly constraining the MLLM’s internal attention impairs its reasoning abilities. We find that $\lambda_{S}=0.01$ strikes the optimal balance and adopt it as our final setting. Notably, this value represents the best result in our discrete ablation set; further fine-grained searches could potentially yield a marginally better optimum. The improvement from this supervision is small on the EarthReason benchmark, which we attribute to its limited scale variation and the predominance of large objects.

\begin{figure}[t]
    \centering
    \includegraphics[width=0.46\textwidth]{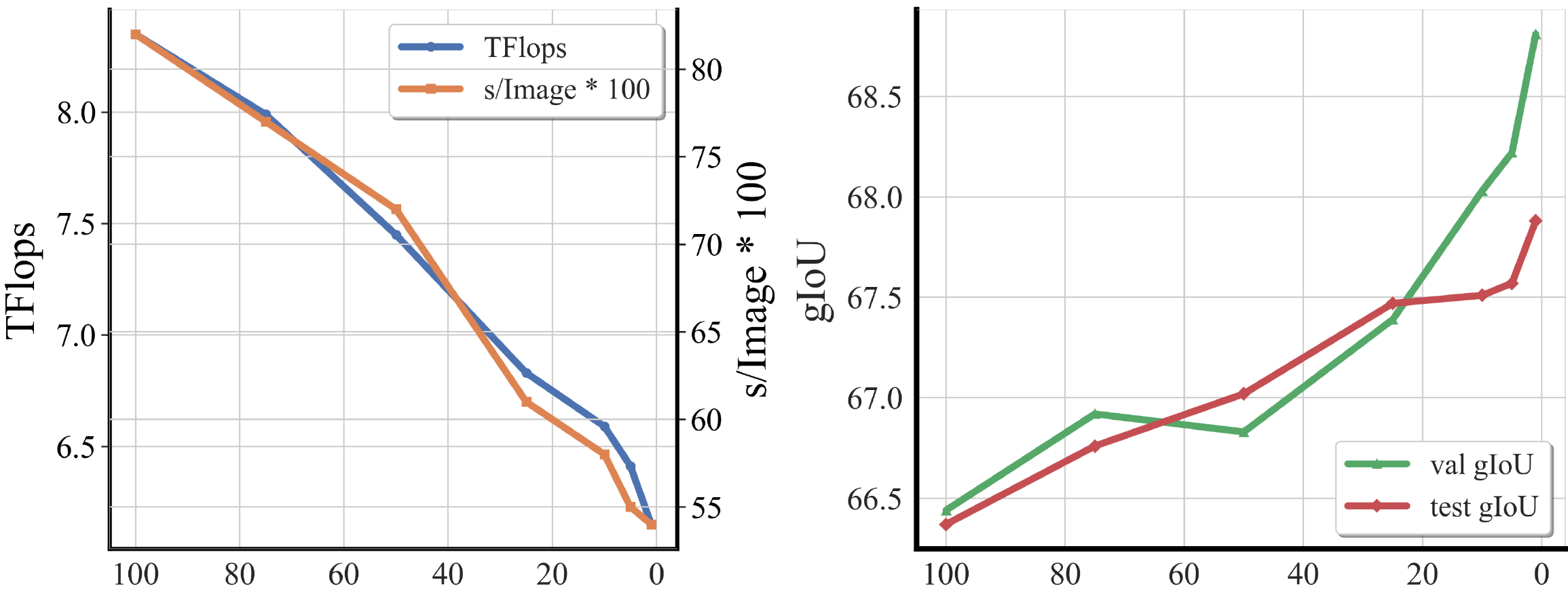}
    \caption{
    The left figure shows that as the number of segmentation queries decreases, both computational cost (TFLOPs) and inference time are correspondingly reduced. The right figure illustrates that this reduction in queries leads to a gradual increase in gIoU scores on the RRSIS-D validation and test sets.}
    \label{fig:ablation}
\end{figure}

\begin{table}[t]
    \centering
    \caption{Ablation experiment on $\lambda_{S}$, which controls the strength of the spatial attention supervision in the MLLM inner layers. The metric is gIoU.}
    \scalebox{0.80}{



\begin{tabular}{c|c|cc|cc|cc}
\hline
\multirow{2}{*}{$\lambda_S$} & \multirow{2}{*}{$\mathcal{L}_{S}$} & \multicolumn{2}{c|}{RRSIS-D} & \multicolumn{2}{c|}{RefSegRS} & \multicolumn{2}{c}{EarthReason}\\ 
& & val & test & val & test & val & test \\ \hline
0 & $\times$ & 68.5 & 66.6 & 81.1 & 70.4 & 72.2 & 72.9 \\
0.1 & \checkmark & 68.1 & 67.3 & 82.2 & 71.1 & 70.1 & 71.8 \\
0.05 & \checkmark & 68.6 & 67.7 & 82.2 & 71.7 & 72.1 & 73.0 \\
\rowcolor{linegray}
0.01 & \checkmark & \textbf{68.8} & \textbf{67.9} & \textbf{82.4} & \textbf{71.8} & \textbf{72.3} & \textbf{73.5} \\ \hline
\end{tabular}}
    \label{tab:ablation_1}
\end{table} 

\begin{table}[t]
    \centering
    \caption{The comparison between different combination of segmentation head. ViT-H: Vision Transformer Huge, Swin-B: Swin Transformer Base, SAM: Segment Anything Model, M2F: Mask2Former. The metric is gIoU.}
    \scalebox{0.8}{

\begin{tabular}{c|c|cc|cc}
    \hline
    \multirow{2}{*}{Mask Decoder} & \multirow{2}{*}{Visual Encoder} & \multicolumn{2}{c|}{EarthReason} & \multicolumn{2}{c}{RefSegRS} \\
    & & val & test & val & test \\ \hline
    SAM~\cite{kirillov2023segment} & ViT-H~\cite{vaswani2017attention} & 62.7 & 65.4 & 73.9 & 60.3 \\
    SAM 2~\cite{ravi2024sam} & ViT-H~\cite{vaswani2017attention} & 64.3 & 64.9 & 74.2 & 64.2 \\
    \rowcolor{linegray}
    M2F~\cite{cheng2022masked} & Swin-B~\cite{liu2021swin} & \textbf{72.3} & \textbf{73.5} & \textbf{82.4} & \textbf{71.8} \\ \hline
\end{tabular}
}
    \label{tab:ablation_3}
\end{table}

\noindent \textbf{Segmentation Query Mechanism Effect.} Previous works~\cite{wei2024instructseg, zhang2024psalm, wei2024hyperseg, wang2025x} use 100 segmentation queries to generate candidate masks and select the best one via a matching algorithm. Our experiments show that this approach is both redundant and inefficient.
Specifically, we evaluate the number of segmentation queries $k$, using values $k \in \{100, 75, 50, 25, 10, 3, 1\}$. When $k > 1$, the single [SEG] token is replaced by a sequence of $k$ independent query tokens, [SEG1] [SEG2] ... [SEG$k$]. The model correspondingly generates $k$ proposal masks, and a matching algorithm~\cite{cheng2022masked} is then employed to select the optimal mask from the $k$ candidates. When $k=1$, the model generates a single mask that is directly used as the final output, bypassing the matching process. As shown in Figure~\ref{fig:ablation}, reducing the number of segmentation queries from 100 to just 1 results in a 34.1\% reduction in inference time (from 0.82 to 0.54 s/image) and a 27.4\% decrease in computational cost (from 8.4 to 6.1 TFLOPs), while also improving the gIoU score on the RRSIS-D~\cite{liu2024rotated} validation set by +2.4 points (66.4 $\rightarrow$ 68.8) and on the test set by +1.6 points (66.3 $\rightarrow$ 67.9). Visual analysis is provided in the Suppl.~\ref{sec:seg_query}.

\noindent \textbf{Combination of Segmentation Head Analysis.} We substitute the Pixel Decoder and Transformer Decoder module (Figure~\ref{fig:overview}) with SAM or SAM2 decoder. Concurrently, the Hierarchical Visual Encoder (Swin-B) is also replaced with the corresponding ViT-H backbone. Table~\ref{tab:ablation_3} shows the results on EarthReason~\cite{li2025segearth} and RefSegRS~\cite{yuan2024rrsis}. The combination of a multi-scale Swin Transformer and a Mask2Former decoder significantly outperforms architectures based on ViT-H and SAM or SAM2 decoder. This confirms that a hierarchical vision backbone, which is adept at capturing features across a wide range of scales, is better suited for the unique challenges of RS images compared to monolithic vision transformers. Our architectural choices are thus empirically validated as optimal for RS domain.

\section{Conclusion}
In this work, we explore comprehensive language-guided segmentation in RS. To advance this task, we present LaSeRS, the first large-scale benchmark designed to systematically investigate four key dimensions of language-guided segmentation: hierarchical granularity, target multiplicity, complex reasoning, and linguistic variability. This benchmark provides the community with a valuable resource to drive progress beyond simple, single-target instructions.
To address the challenges of this benchmark, we propose SegEarth-R2, an effective and efficient MLLM architecture. Our evaluations show that SegEarth-R2 not only excels on the challenges posed by LaSeRS but also establishes a new SOTA across a range of established referring and reasoning segmentation benchmarks. Ultimately, we hope this work catalyzes a paradigm shift in the RS community, moving beyond parsing simple, single-target instructions toward developing comprehensive, unified language-guided segmentation.
{
    \small
    \bibliographystyle{ieeenat_fullname}
    \bibliography{main}
}

\clearpage
\setcounter{page}{1}
\maketitlesupplementary

\section{More information of LaSeRS}
\label{sec:more_information}
\subsection{Definition of Four Dimensions of LaSeRS}
\noindent \textbf{\textcolor[RGB]{45,83,150}{Hierarchical Granularity.}}
A primary challenge in RS arises from the vast and complex nature of the imagery. A single scene often contains a diverse array of objects at vastly different scales, from large-scale collections (\eg, ``all indistinguishable vehicles densely parked in the parking lot") to specific instances (\eg, ``a cargo truck parked in the bottom left"). Consequently, language queries used to describe these scenes inherently exhibit hierarchical granularity. As shown in Figure~\ref{fig:examples}, this hierarchical granularity can be categorized into conceptual and segmentation granularity. The first, hierarchical concept granularity, relates to the level of semantic abstraction in the query, requiring the model to understand a hierarchy from coarse categories (\eg, ``airplane") to fine-grained sub-classes (\eg, ``Boeing 787"). This is particularly challenging in RS images due to noise, small objects, and the need for expert knowledge. The second, hierarchical segmentation granularity, dictates the spatial scope of the desired output mask, which operates at three distinct levels: semantic (\eg, ``all tennis courts"), instance (\eg, ``the rightmost tennis court"), and part (\eg, ``the service area of the rightmost tennis court").

\noindent \textbf{\textcolor[RGB]{83,129,52}{Target Multiplicity.}}
This challenge arises from queries that reference multiple objects simultaneously, testing a model's ability to parse and ground complex instructions. This is illustrated by the target multiplicity and long query cases in Figure~\ref{fig:examples}.

\noindent \textbf{\textcolor[RGB]{187,137,0}{Reasoning Requirements.}}
This challenge spans a spectrum from explicit to implicit reasoning~\cite{li2025segearth}. Explicit queries relate to literal visual attributes (\eg, ``a large ground track field"). In contrast, implicit queries require commonsense knowledge; for example, processing ``an escape direction in case of an earthquake" requires the model to infer safety by identifying open areas and avoiding dense structures.

\noindent \textbf{\textcolor[RGB]{149,0,149}{Linguistic Variability.}} As shown in Figure~\ref{fig:examples}, this incorporates both concise and long, descriptive queries to evaluate model robustness to varying levels of linguistic detail.

\subsection{The Details of Masks Generation}
As shown in Figure~\ref{fig:pipeline}, we employ two distinct point prompting strategies to generate candidate masks. For global point prompts, we set a grid parameter $R=C=4$, uniformly sampling an $R \times C$ grid of 16 points across the square image to prompt the SAM. For local point prompts, we first utilize the provided bbox annotation to crop the target region. Points are then sampled exclusively within this cropped area. Assuming the cropped region has a height of $h$ and a width of $w$, the sampling grid dimensions $(R, C)$ are determined by:
\begin{equation*}
\resizebox{1.0\linewidth}{!}{$
    (R, C) =
    \begin{cases}
        \left( \left\lceil \frac{\max\{h, w\}}{\min\{h, w\}} \right\rceil + 1, 1 \right), & \text{if } \frac{\max\{h, w\}}{\min\{h, w\}} \ge 2.5 \\
        \\
        (4, 4), & \text{if } 1 \le \frac{\max\{h, w\}}{\min\{h, w\}} < 2.5
    \end{cases}
$}
\end{equation*}

\subsection{The Details of Masks Filtering}
\label{sec:mask_filtering}
In this section, we detail the automated mask filtering pipeline, which is visually depicted in Figure~\ref{fig:pipeline}. This process is designed to programmatically curate the dataset by removing low-quality or erroneous annotations. We use the ``airplane" category as an illustrative example to describe this multi-stage procedure.

First, we perform a coarse-grained sanity check based on object counts. We iterate through all samples and discard any instance where the number of discrete connected components in the binary mask does not precisely match the number of associated bounding boxes. This step effectively removes clear anomalies, such as fragmented masks (\ie, multiple components for one box) or improperly merged masks (\ie, one component for multiple boxes).

Second, from the pool of samples that pass this initial check, we establish a ``gold standard" reference set. This set consists of 50 high-quality ``airplane" masks that were manually selected and verified by human annotators. This reference set serves as the exemplar distribution from which we derive the target geometric profile for the category.

Third, we derive and apply a set of heuristic thresholds by statistically analyzing this ``gold standard" set. We calculate the following geometric properties for the primary connected component of each of the 50 reference masks. All calculations are implemented using the OpenCV library.

\begin{itemize}
\item \textbf{Eccentricity}: Measures how much the shape of the mask deviates from a perfect circle.
\item \textbf{Circularity}: Quantifies the ``roundness" of the mask, defined as $\frac{4\pi \cdot \text{Area}}{\text{Perimeter}^2}$.
\item \textbf{Solidity}: The ratio of the mask's area to the area of its convex hull. This metric penalizes shapes with significant indentations.
\item \textbf{Symmetry}: A custom metric defined to quantify the object's expected bilateral symmetry.
\item \textbf{Extent}: The ratio of the mask's area to the area of its bounding box ($\frac{\text{Mask Area}}{\text{Bbox Area}}$). This filters masks that are overly sparse.
\end{itemize}

By calculating the distribution (\eg, mean and standard deviation) of these properties across the reference set, we establish an ``acceptable range" for each metric. Finally, these ranges are applied as a fine-grained filter to all remaining masks in the dataset. A mask is retained only if all its geometric properties fall within these predefined bounds.

\subsection{The Details of QAs Generation}
In this section, we detail the QA generation process, as shown in Figure~\ref{fig:pipeline}. Our methodology adapts based on the mask source, which falls into two categories: those with pre-existing category labels and those without.

\begin{figure*}[h]
    \centering
    \includegraphics[width=0.95\textwidth]
    {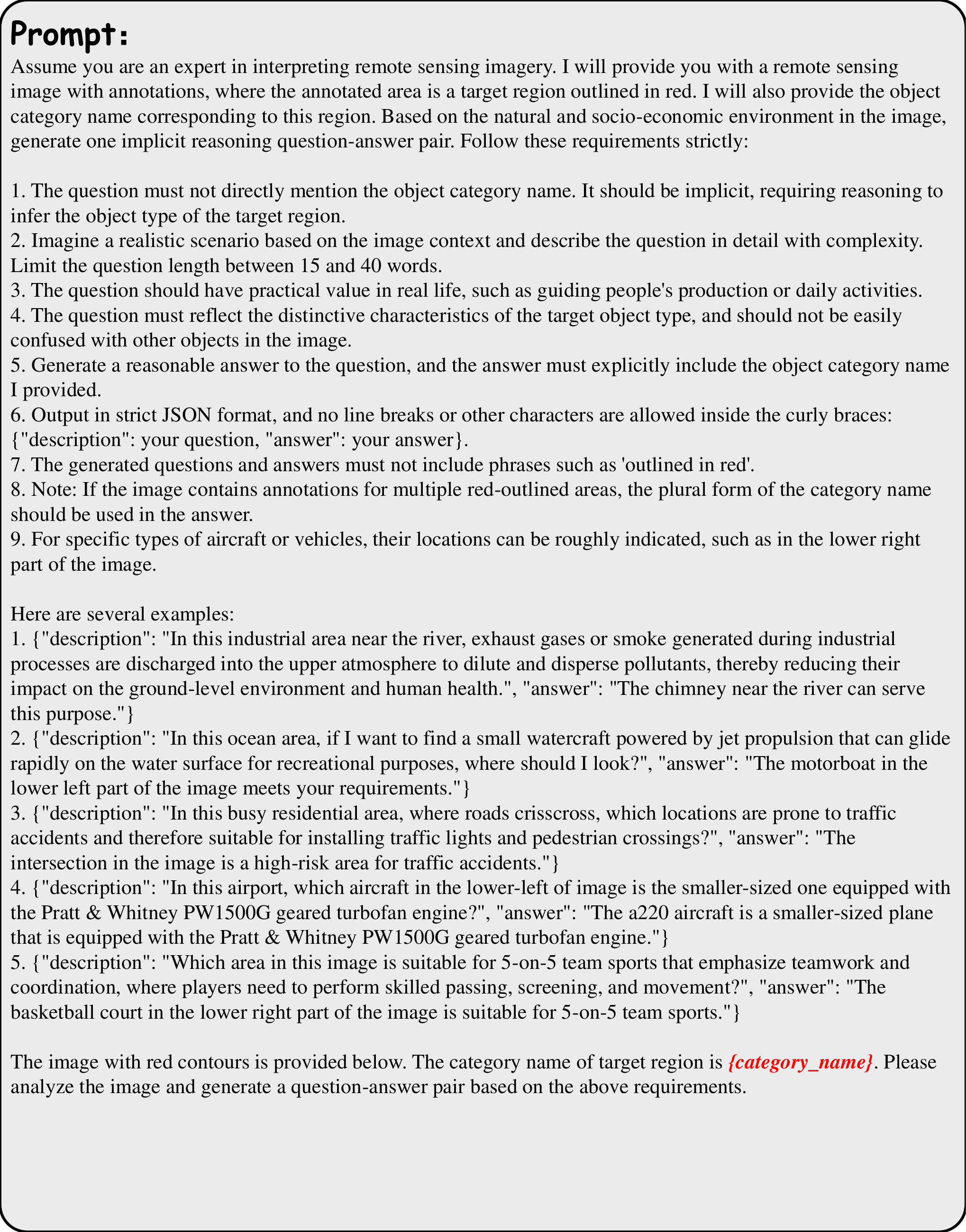}
    \caption{An example of single-target QA generation prompt}
    \label{fig:prompt_1}
\end{figure*}

\begin{figure*}[h]
    \centering
    \includegraphics[width=0.95\textwidth]
    {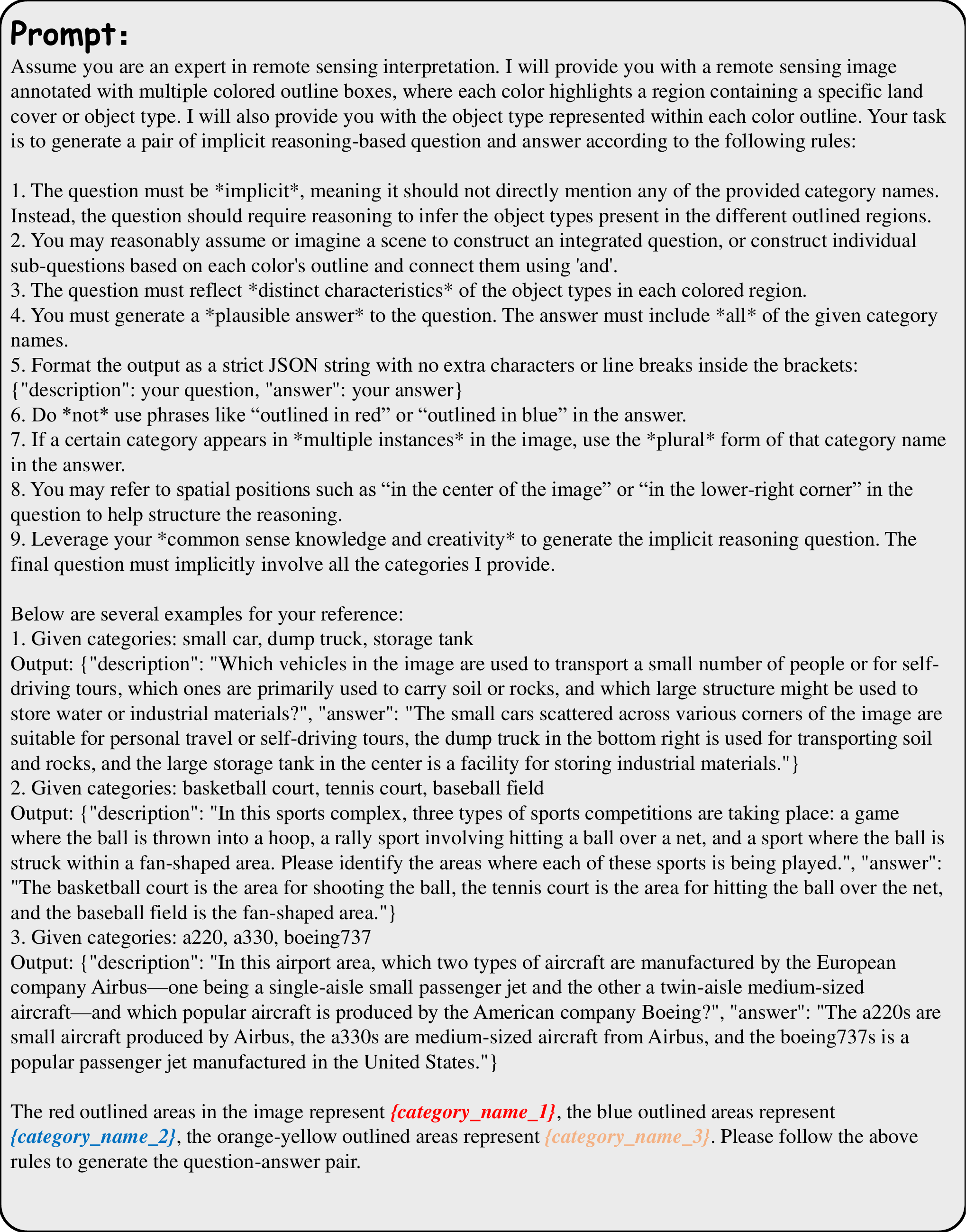}
    \caption{An example of multi-target QA generation prompt}
    \label{fig:prompt_2}
\end{figure*}

For masks with category labels, we employ two prompts, Prompt~\ref{fig:prompt_1} and Prompt~\ref{fig:prompt_2}. Simple adjustments to these prompts generate a diverse set of QA pairs, allowing us to control factors such as output length and implicit reasoning.

\begin{figure*}[h]
    \centering
    \includegraphics[width=0.95\textwidth]
    {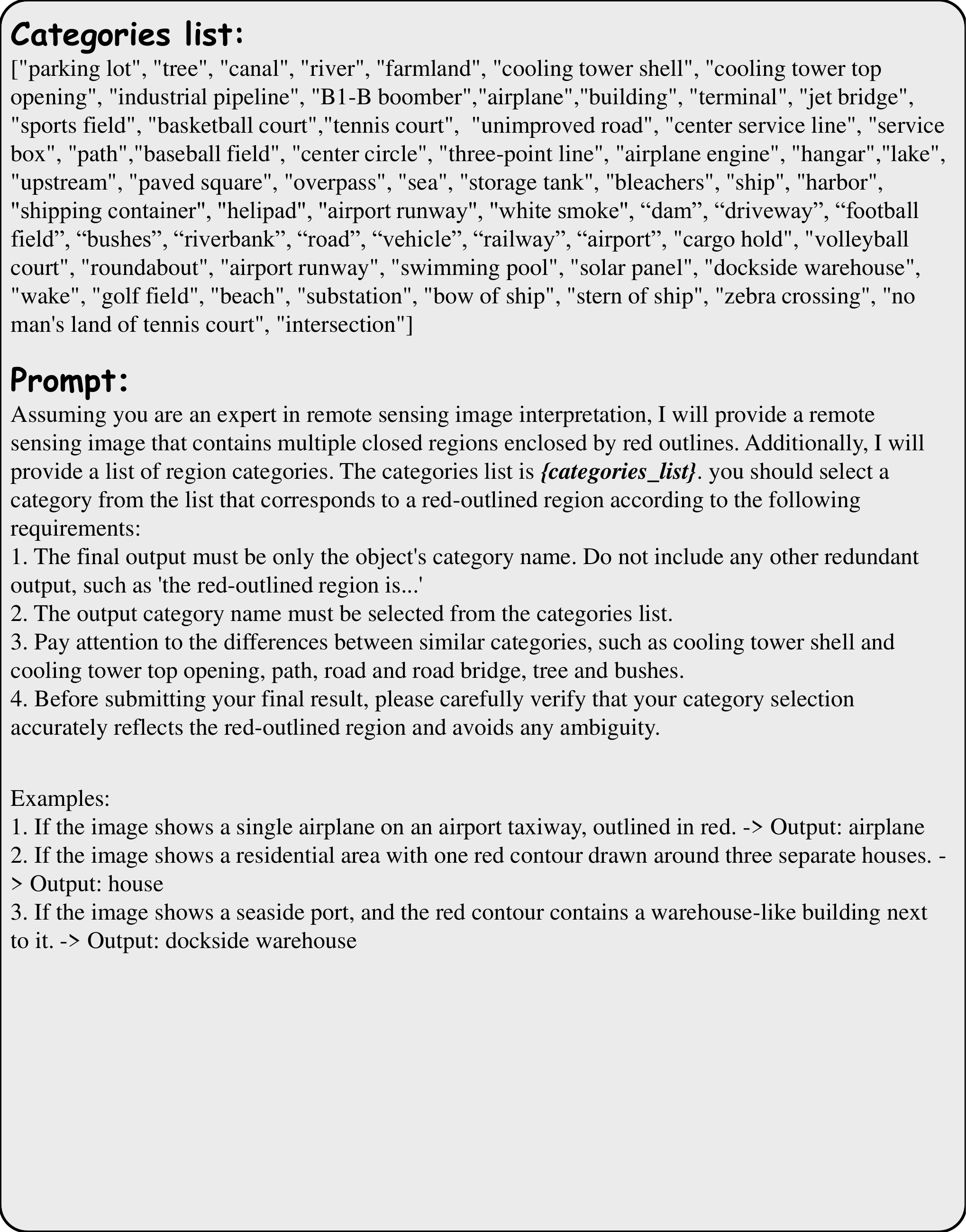}
    \caption{Category generation prompt}
    \label{fig:prompt_3}
\end{figure*}

Conversely, for masks that lack category labels, we first perform a preliminary category generation step. We use Prompt~\ref{fig:prompt_3} to assign a category label to the masked region. Once the label is obtained, we then proceed to generate the corresponding QA pairs using the methods described above.

To execute this entire generation pipeline, we utilize the Gemini-2.5-Pro model, which we selected for its exceptional instruction-following capabilities. To ensure a high degree of diversity in the generated questions, we set the temperature parameter to 1.0. Furthermore, we enable its chain-of-thought reasoning capabilities to maintain high fidelity and coherence in the output.

\subsection{The Details of QAs Filtering}
The initial generation phases produced a raw dataset of approximately 40k QA pairs. To ensure the highest standards of quality and eliminate any generation artifacts, we implemented a rigorous, multi-stage manual validation process. This critical task was performed by a team of 15 domain experts, all with extensive experience in remote sensing and computer vision. 

Each of the 40k QA pairs was meticulously inspected against a four-point quality rubric. A pair was only retained if it passed all four criteria:

\begin{itemize}
    \item \textbf{Object recognition:} Verifying the accuracy of object category recognition in the image.
    \item \textbf{Spatial description and logical consistency of the question:} Ensuring the spatial and logical accuracy of the question’s description.
    \item \textbf{Mask quality:} Assessing if the annotated mask meets quality standards for precision and completeness.
    \item \textbf{Grammatical accuracy:} Checking for grammatical mistakes or inconsistencies in the text.
\end{itemize}

Pairs that failed any of these checks were discarded. Following this comprehensive review, we performed a final quality assurance pass that involved randomly sampling the filtered set to verify the consistency and quality of the experts' work.

This expert-driven curation process yielded our final dataset, comprising 30,830 high-quality QA pairs that, in total, correspond to over 40,000 validated object masks.

\subsection{Category List of LaSeRS}
Table~\ref{tab:categories} illustrates the category distribution of the LaSeRS dataset. Our dataset includes not only general categories but also fine-grained concepts and part-level categories.

\label{sec:category}
\begin{table*}[htbp]
    \centering
    \caption{Comparison of dataset categories: LaSeRS contains 122 categories, encompassing general, fine-grained, and part-level classes, and covering a wide range of remote sensing scenes such as land cover, vehicles, and natural landscapes. In contrast, RefSegRS, RRSIS-D, RISBench, and EarthReason contain only a dozen to around twenty categories.}
    \vspace{-0.5em}
    \scalebox{0.97}
    {\begin{tabularx}{\textwidth}{c|X}
\hline
\textbf{Dataset} & \multicolumn{1}{c}{\textbf{Categories}} \\
\hline
LaSeRS           & \textbf{General}: \textit{airplane, airport, airport runway, bare land, baseball diamond, baseball field, basketball court, beach, bridge, bridge road, building, bushes, canal, chimney, cooling tower, dam, expressway service area, expressway toll station, farmland, football field, golf field, grass, green strip, greenhouse, ground track field, harbor, helicopter, helipad, intersection, jet bridge, lake, large vehicle, overpass, parking lot, path, paved road, paved square, plane, playground, railway, river, road, roundabout, sea, ship, slide, small car, small vehicle, soccer ball field, solar panel, sports field, stadium, storage tank, substation, swimming pool, tennis court, terminal, train station, tree, unimproved road, vehicle, volleyball court, water, white smoke, windmill.}
                   \newline 
                   \textbf{Fine-grained Concept}: \textit{B1-B boomber, a220, a321, a330, a350, arj21, boeing737, boeing747, boeing777, boeing787, bus, c919, cargo truck, container crane, dry cargo ship, dockside warehouse, dump truck, driveway, engineering ship, excavator, fishing boat, hangar, liquid cargo ship, motorboat, passenger ship, tractor, trailer, truck tractor, tugboat, van, warship. 
                   \newline 
                   \textbf{Part}: airplane engine, bleachers, bow of ship, cargo hold, center circle, center line, center service line, cooling tower shell, cooling tower top opening, downstream, football net, fuselage, horizontal stabilizer, industrial pipeline, net, no man's land of tennis court, riverbank, service box, shipping container, stern of ship, tennis net, three-point line, upstream, wake, wing, zebra crossing.} \\
\hline
RefSegRS~\cite{yuan2024rrsis}         & \textbf{General}: \textit{road, vehicle, car, van, building, truck, trailer, bus, road marking, bikeway, sidewalk, tree, low vegetation, impervious surface.} \\
\hline
RRSIS-D~\cite{liu2024rotated}          & \textbf{General}: \textit{airplane, airport, golf field, expressway service area, baseball field, stadium, ground track field, storage tank, basketball court, chimney, tennis court, overpass, train station, ship, express toll station, dam, harbor, bridge, vehicle, windmill.} \\
\hline
RISBench~\cite{dong2024cross} & \textbf{General}: \textit{expressive service area, expressive toll station, ground track field, basketball court, container crane, roundabout, windmill, overpass, stadium, bridge, soccer ball field, baseball diamond, train station, golf field, airport, harbor, dam, ship, helipad, vehicle, chimney, airplane, helicopter, tennis court, storage tank, swimming pool.} \\
\hline
EarthReason~\cite{li2025segearth} & \textbf{General}: \textit{storage tank, bridge, intersection, tennis court, baseball field, substation, pier, viaduct, wind turbine, church, airport runway, swimming pool, lake, airport helipad, dam, railway, basketball court, beach, greenhouse, roundabout, solar power plant, ground track field, waterwaste plant, river, train station, stadium, island, factory.} \\
\hline
\end{tabularx}
}
    \label{tab:categories}
\end{table*}

\subsection{More Examples of LaSeRS}
\begin{figure*}[h]
    \centering
    \includegraphics[width=0.95\textwidth]
    {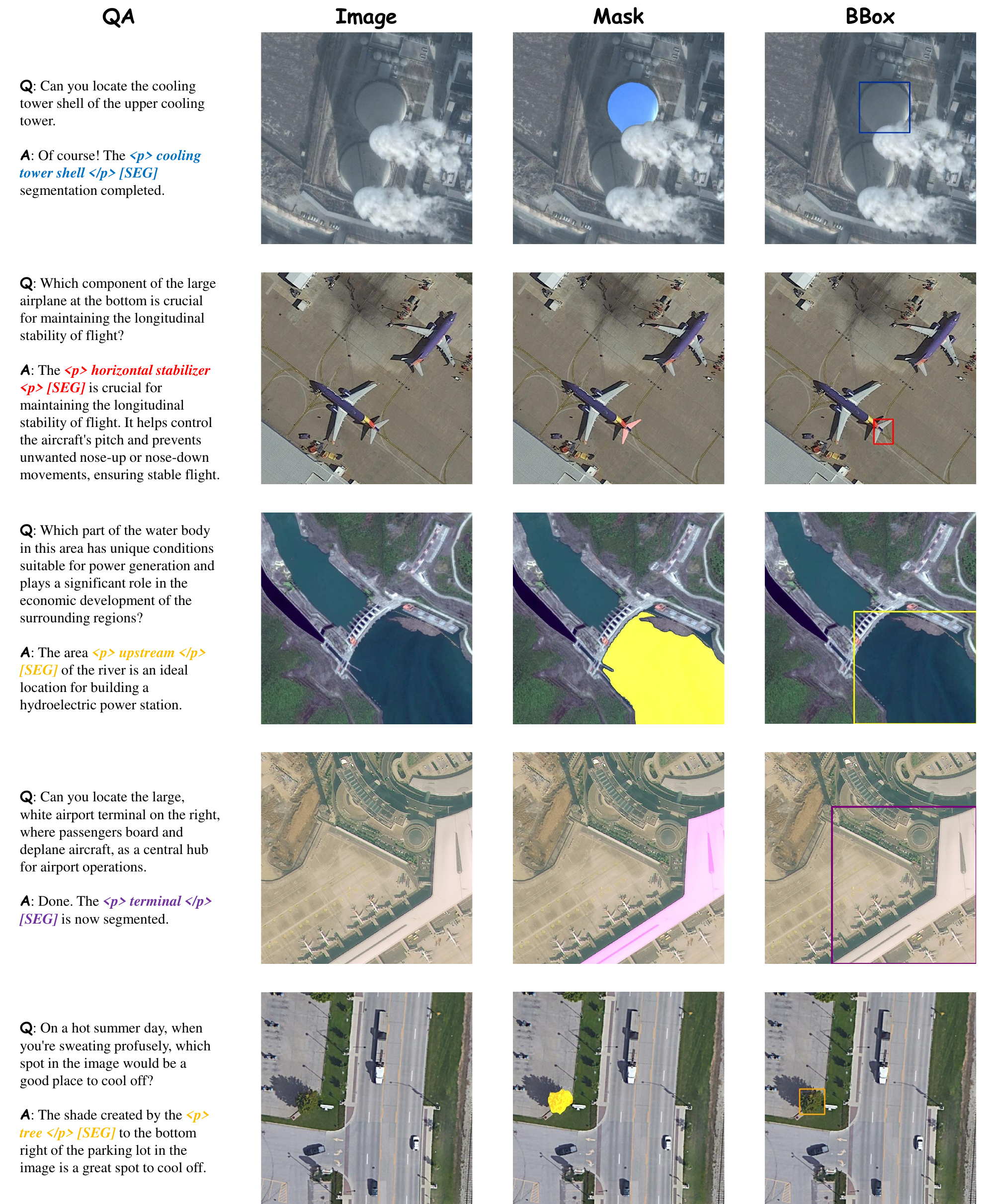}
    \caption{More examples of LaSeRS.}
    \label{fig:appendix_examples_1}
\end{figure*}

Figures~\ref{fig:appendix_examples_1} and~\ref{fig:appendix_examples_2} show more examples of LaSeRS.

\section{Experiment Setting Details}
\subsection{Training Details}
Table~\ref{tab:training} presents the specific training hyperparameters of SegEarth-R2. Training and testing were conducted on an Nvidia A100 80GB GPU.
\label{sec:training_details}
\begin{table*}[t]
    \centering
    \caption{Hyper parameters of our model in the training.}
    \scalebox{1}{\begin{tabular}{lr}
    \toprule
    \textbf{Parameters} & \textbf{Value} \\
    \midrule
    Optimizer & AdamW \\
    Learning Rate & $1 \times 10^{-4}$ \\
    Batch Size & 4 \\
    Training Steps & 50,000 \\
    Learning Rate Schedule & Cosine Decay \\
    Lora Rank & 8 \\
    Computation Precision & bf16 \\
    Weight Decay & 0.0 \\
    Warmup Ratio & 0.03 \\
    Image Size & $1024 \times 1024$ \\
    Image Processing & \begin{tabular}[t]{@{}r@{}} 
                           Resize long edge to 1024 \\
                           and padding short edge to 1024.
                         \end{tabular} \\
    \bottomrule
\end{tabular}}
    \label{tab:training}
\end{table*}

\section{Visual Analysis}
\subsection{Visual Analysis of Spatial Attention}
\label{sec:spatial_attention}
\begin{figure*}[h]
    \centering
    \includegraphics[width=0.95\textwidth]
    {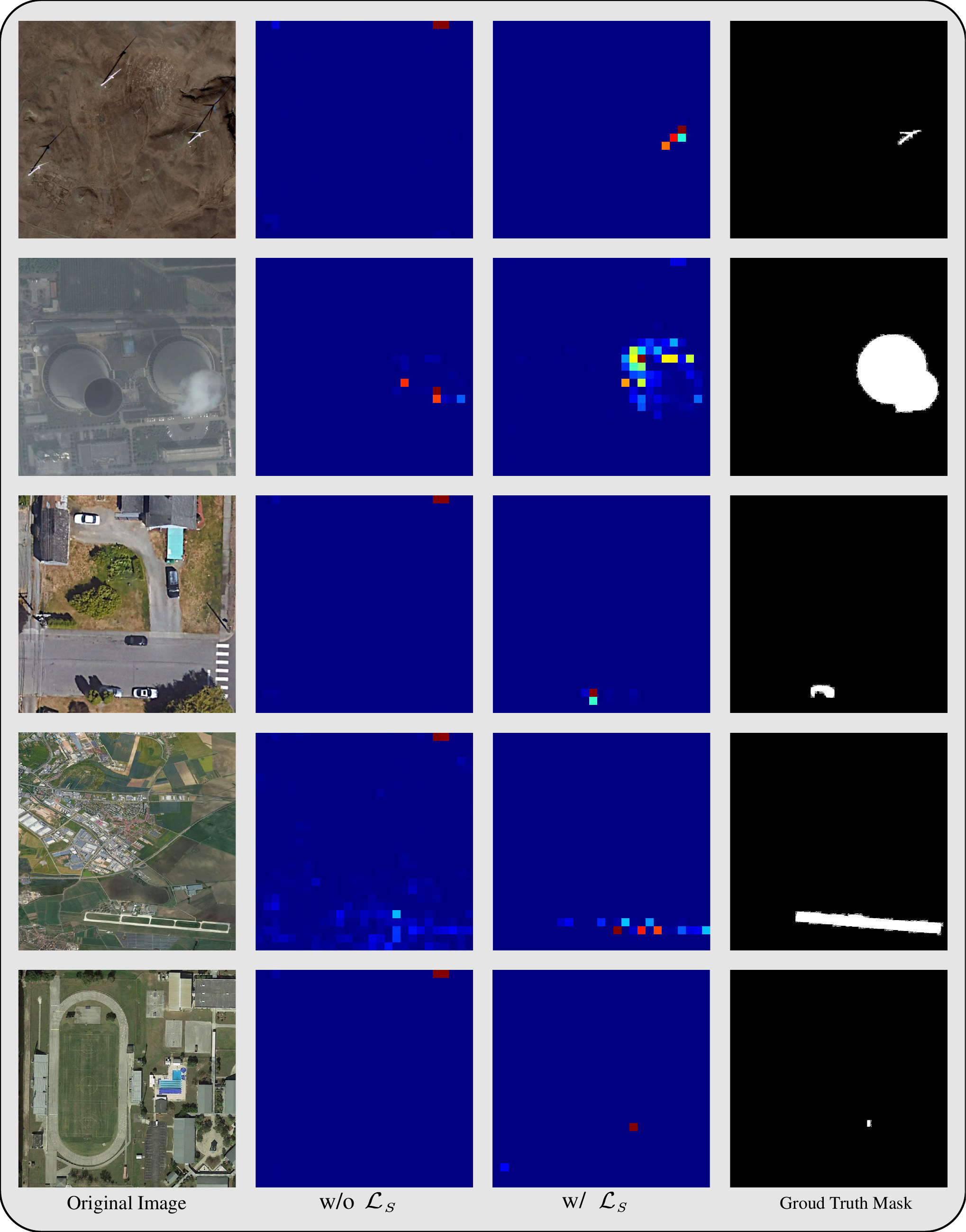}
    \caption{Additional examples of spatial attention visualizations.}
    \label{fig:visual_analysis}
\end{figure*}

Figure~\ref{fig:visual_analysis} shows additional visualizations of the spatial attention mechanism. It can be observed that our spatial attention supervision helps the model better localize complex or small objects.
\subsection{Visual Analysis of Segmentation Query}
\label{sec:seg_query}
\begin{figure*}[h]
    \centering
    \includegraphics[width=1.0\textwidth]
    {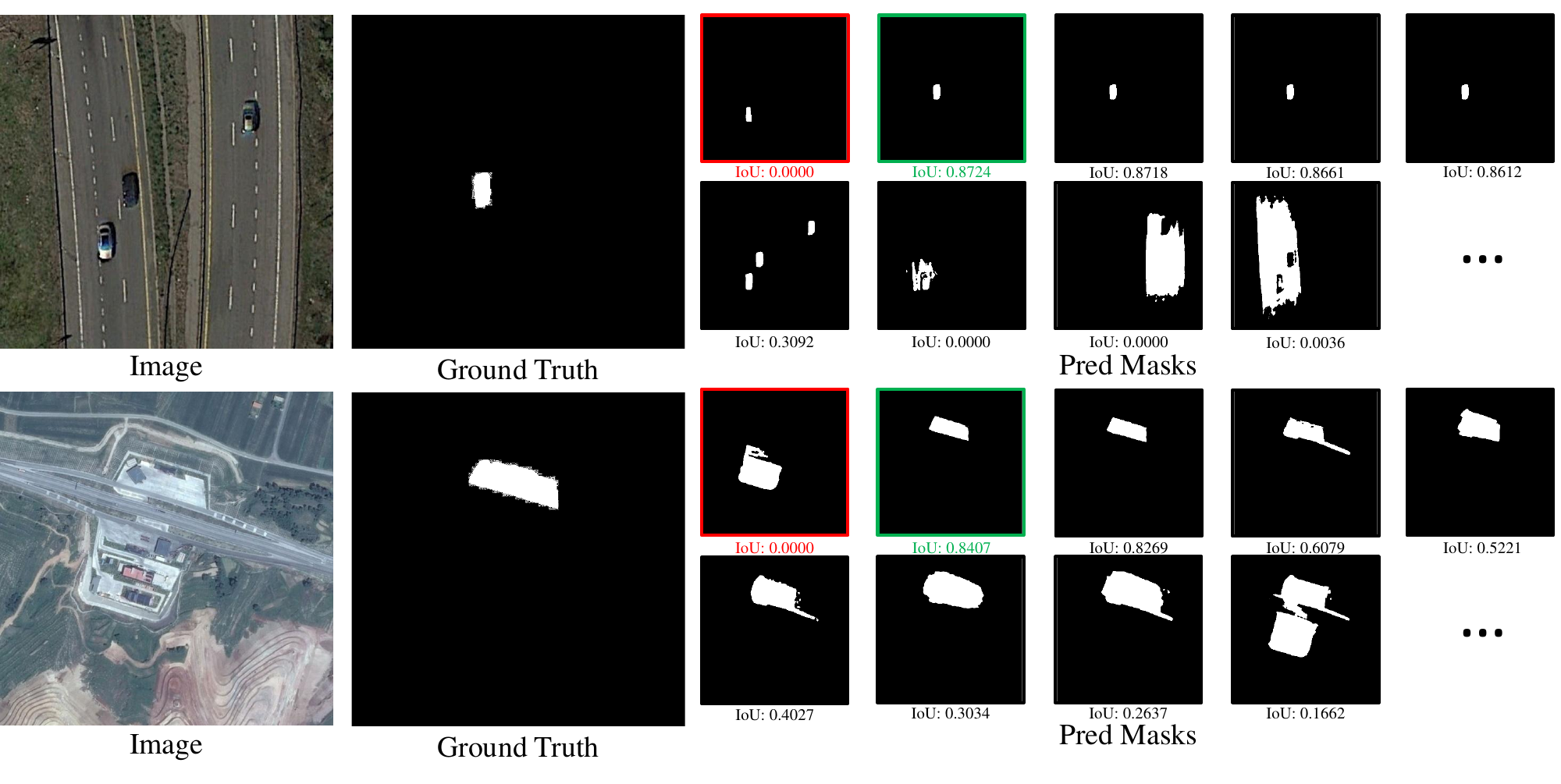}
    \caption{Visualization analysis of segmentation query number. The \textcolor{myred}{red} box indicates the mask selected by the matching algorithm, while the \textcolor{mygreen}{green} shows the mask that best matches the ground truth. It can be observed that the matching algorithm always selects a non-optimal mask, leading to a decrease in both efficiency and performance.}
    \label{fig:seg_query_1}
\end{figure*}

\begin{figure*}[h]
    \centering
    \includegraphics[width=1.0\textwidth]
    {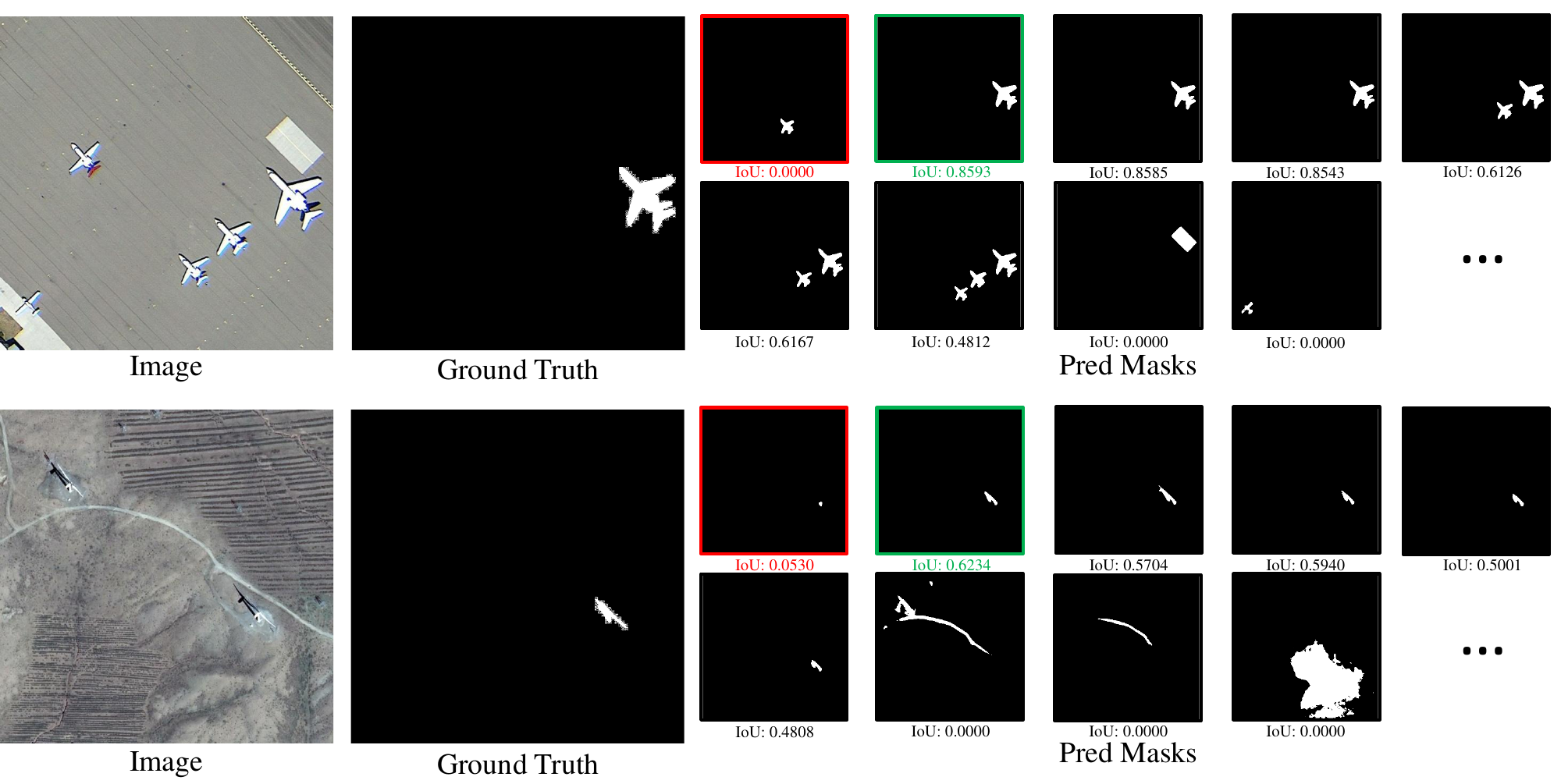}
    \caption{Visualization analysis of segmentation query number. The \textcolor{myred}{red} box indicates the mask selected by the matching algorithm, while the \textcolor{mygreen}{green} shows the mask that best matches the ground truth. It can be observed that the matching algorithm always selects a non-optimal mask, leading to a decrease in both efficiency and performance.}
    \label{fig:seg_query_2}
\end{figure*}

\begin{figure*}[h]
    \centering
    \includegraphics[width=1.0\textwidth]
    {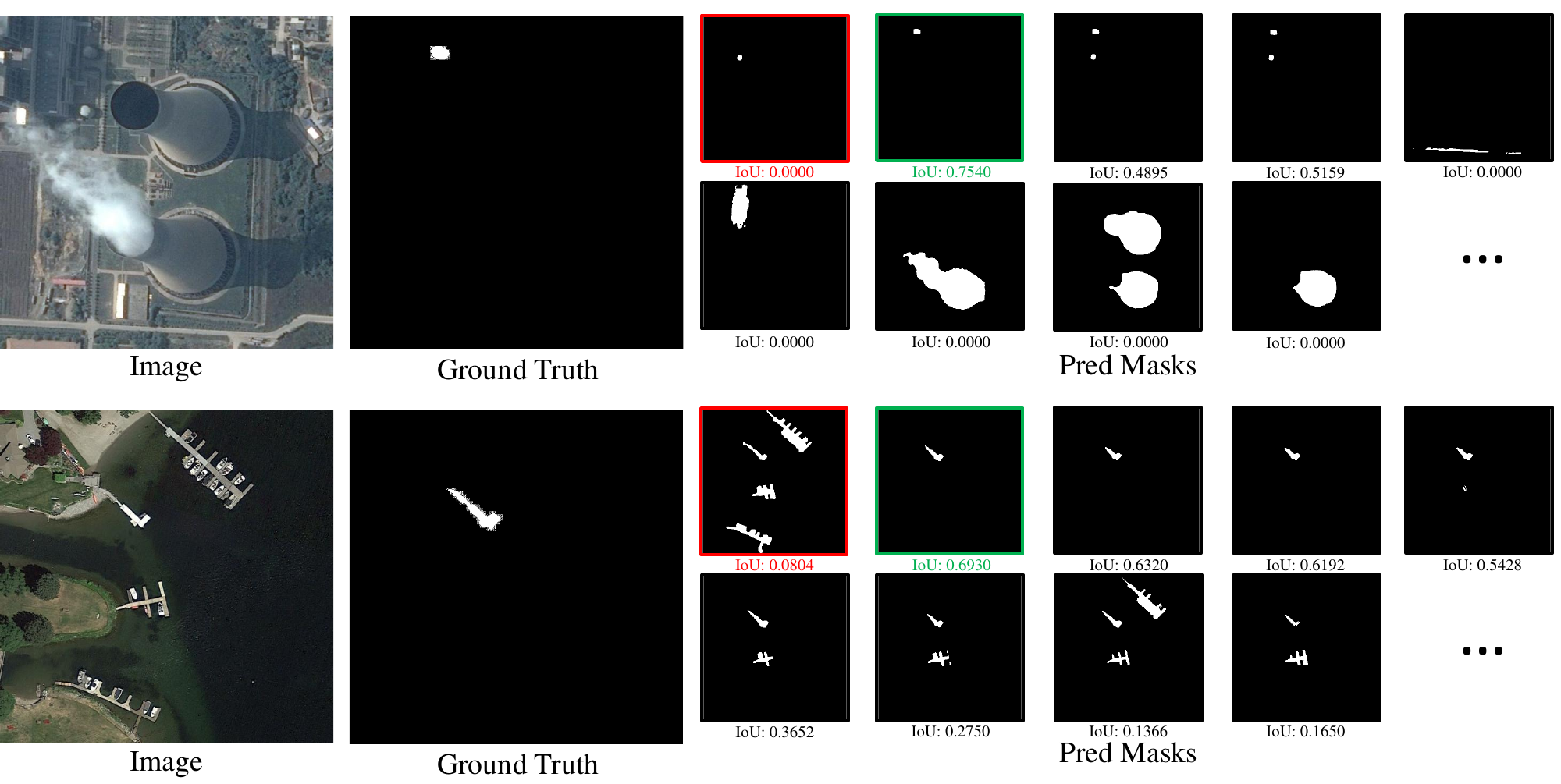}
    \caption{Visualization analysis of segmentation query number. The \textcolor{myred}{red} box indicates the mask selected by the matching algorithm, while the \textcolor{mygreen}{green} shows the mask that best matches the ground truth. It can be observed that the matching algorithm always selects a non-optimal mask, leading to a decrease in both efficiency and performance.}
    \label{fig:seg_query_3}
\end{figure*}

\begin{figure*}[h]
    \centering
    \includegraphics[width=1.0\textwidth]
    {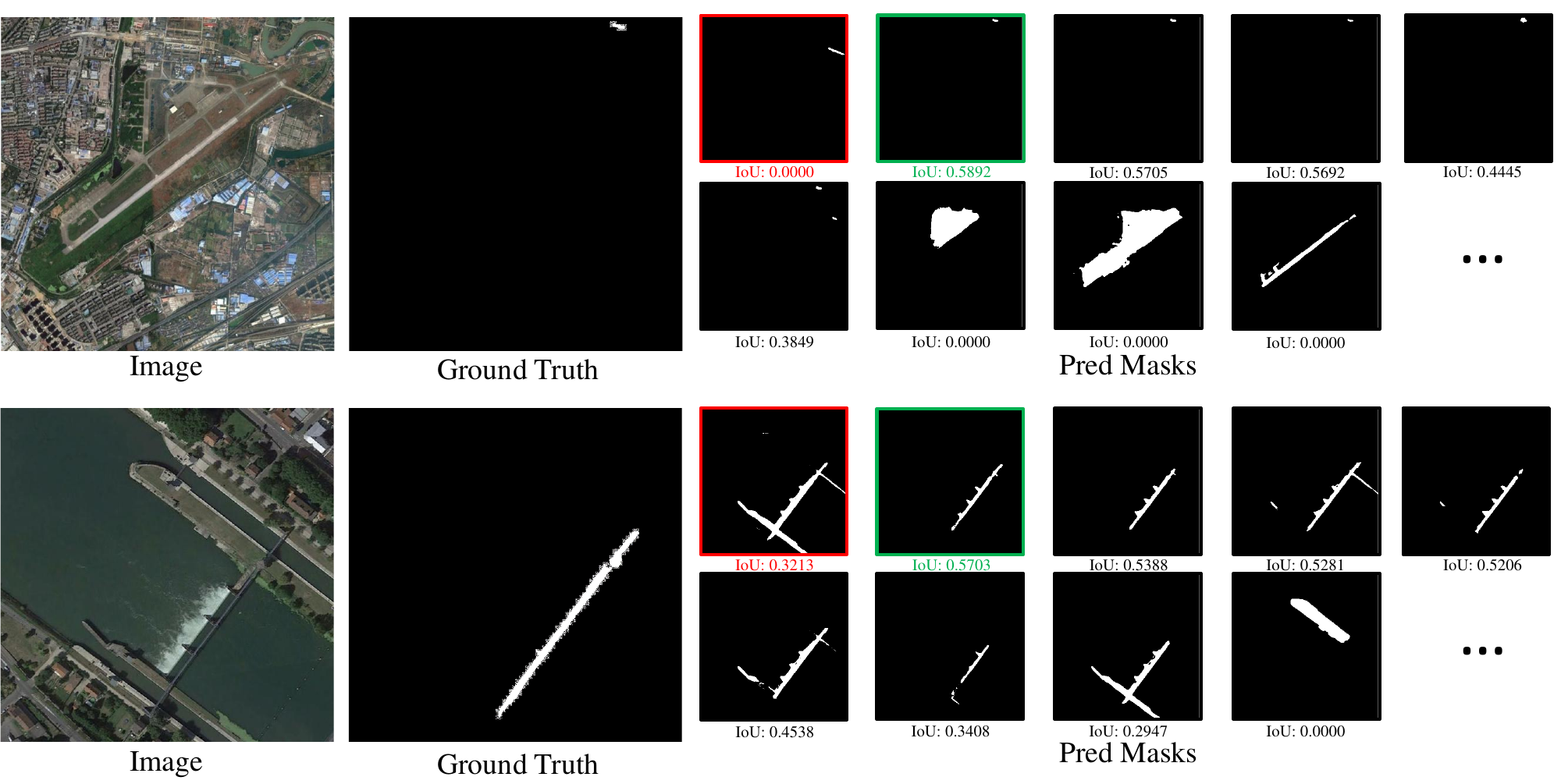}
    \caption{Visualization analysis of segmentation query number. The \textcolor{myred}{red} box indicates the mask selected by the matching algorithm, while the \textcolor{mygreen}{green} shows the mask that best matches the ground truth. It can be observed that the matching algorithm always selects a non-optimal mask, leading to a decrease in both efficiency and performance.}
    \label{fig:seg_query_4}
\end{figure*}

Figure~\ref{fig:seg_query_1}, Figure~\ref{fig:seg_query_2}, Figure~\ref{fig:seg_query_3} and Figure~\ref{fig:seg_query_4} show the ground truth mask, the mask selected by matching algorithm and the candidate masks when segmentation query number is set to 100. In many cases, the mask selected by the matching algorithm is not optimal, which leads to losses in both efficiency and performance. We set the segmentation query number to 1, and Figure~\ref{fig:ablation} demonstrates the results. It can be observed that both efficiency and performance have been significantly improved.

\section{Qualitative Results of SegEarth-R2}

\subsection{LaSeRS}
\label{sec:LaSeRS}
\begin{figure*}[h]
    \centering
    \includegraphics[width=0.95\textwidth]
    {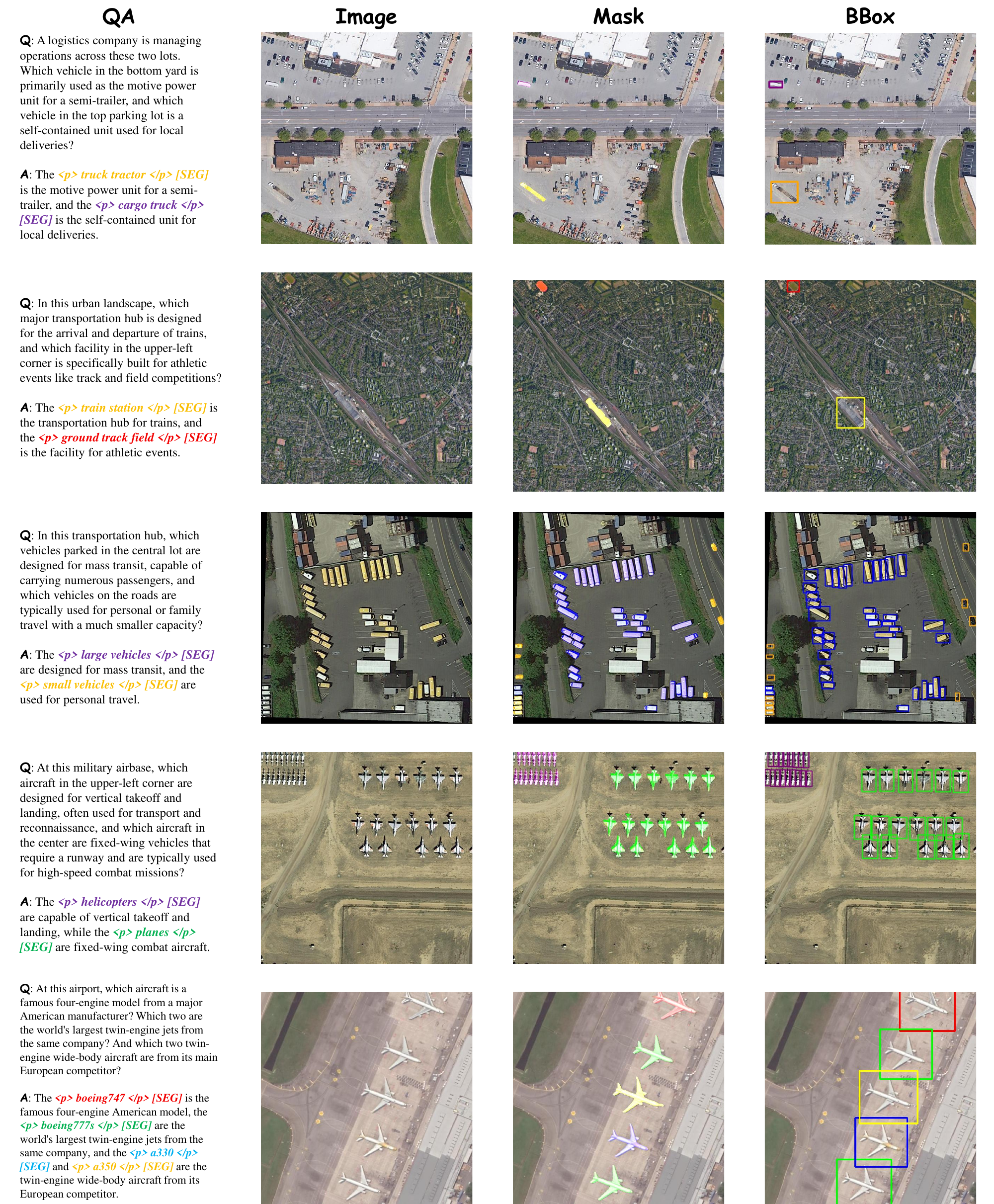}
    \caption{More examples of LaSeRS.}
    \label{fig:appendix_examples_2}
\end{figure*}

\begin{figure*}[h]
    \centering
    \includegraphics[width=0.95\textwidth]
    {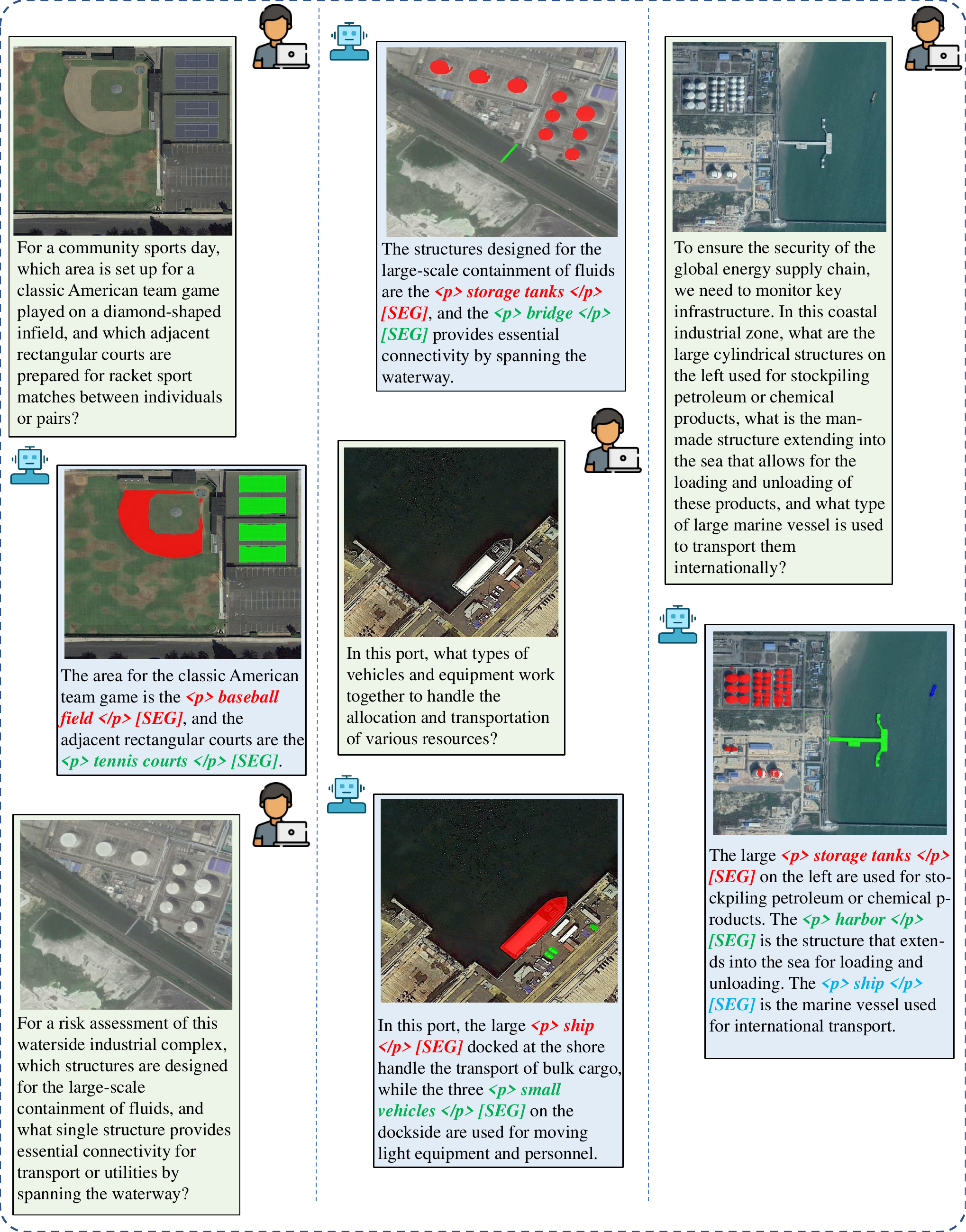}
    \caption{Visualization results of SegEarth-R2 on the LaSeRS dataset.}
    \label{fig:qualitative_results_1}
\end{figure*}

Figure~\ref{fig:qualitative_results_1} and Figure~ \ref{fig:appendix_examples_2} show the qualitative results on the LaSeRS test set.

\subsection{RefSegRS, RRSIS-D, and RISBench}
\label{sec:referring}
\begin{figure*}[h]
    \centering
    \includegraphics[width=0.95\textwidth]
    {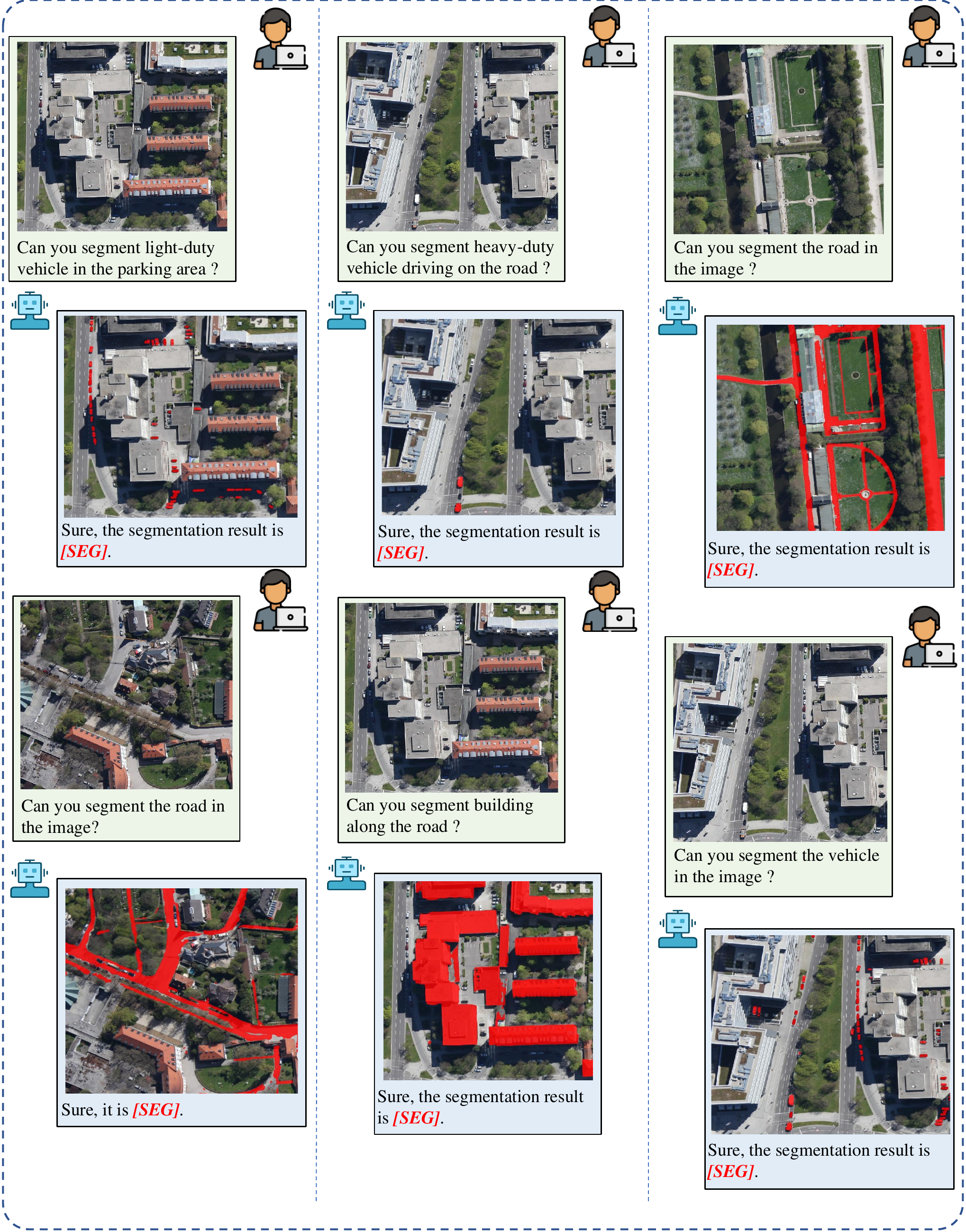}
    \caption{Visualization results of SegEarth-R2 on the RefSegRS test set.}
    \label{fig:qualitative_results_refsegrs}
\end{figure*}

\begin{figure*}[h]
    \centering
    \includegraphics[width=0.95\textwidth]
    {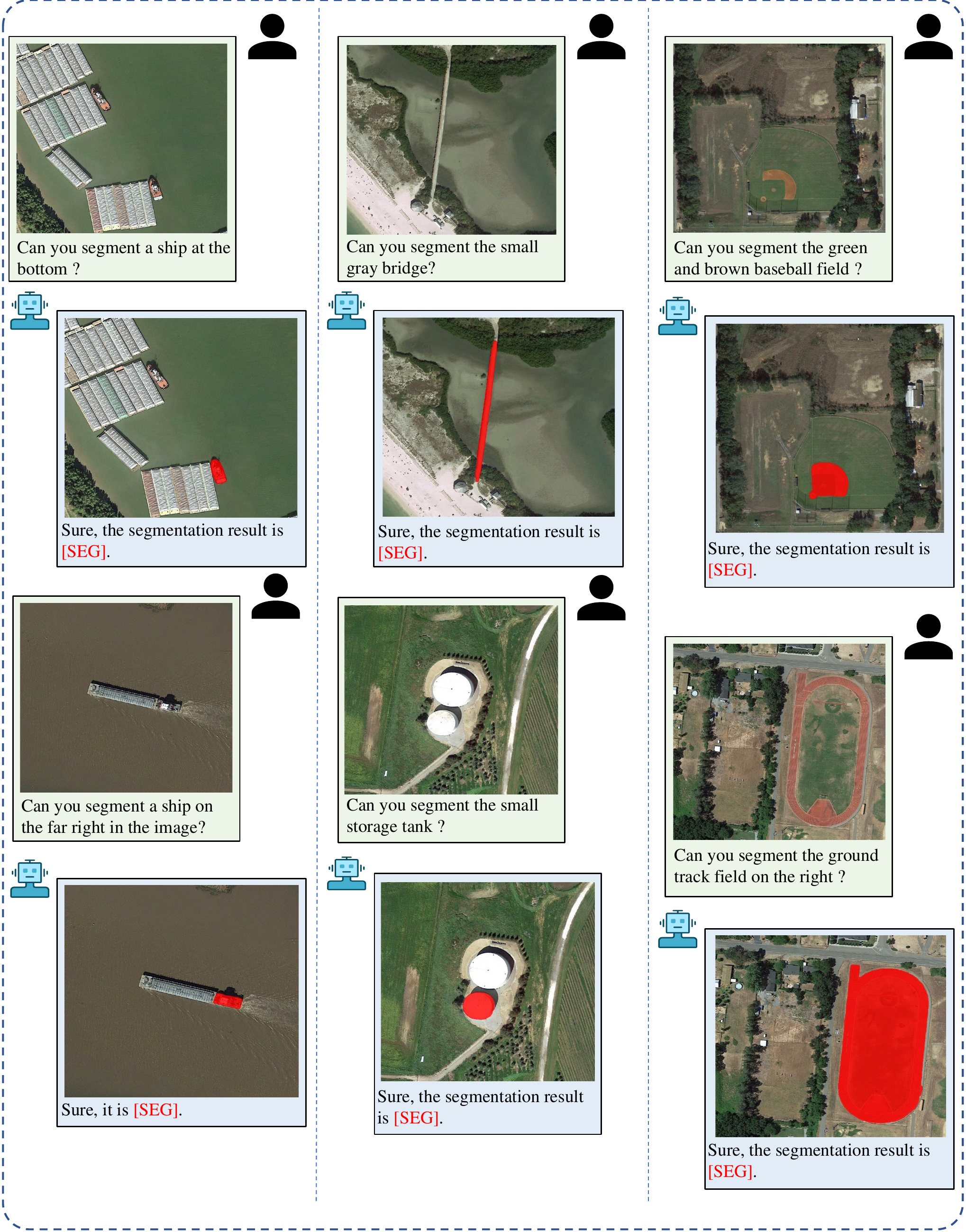}
    \caption{Visualization results of SegEarth-R2 on the RRSIS-D test set.}
    \label{fig:qualitative_results_rrsis-d}
\end{figure*}

\begin{figure*}[h]
    \centering
    \includegraphics[width=0.95\textwidth]
    {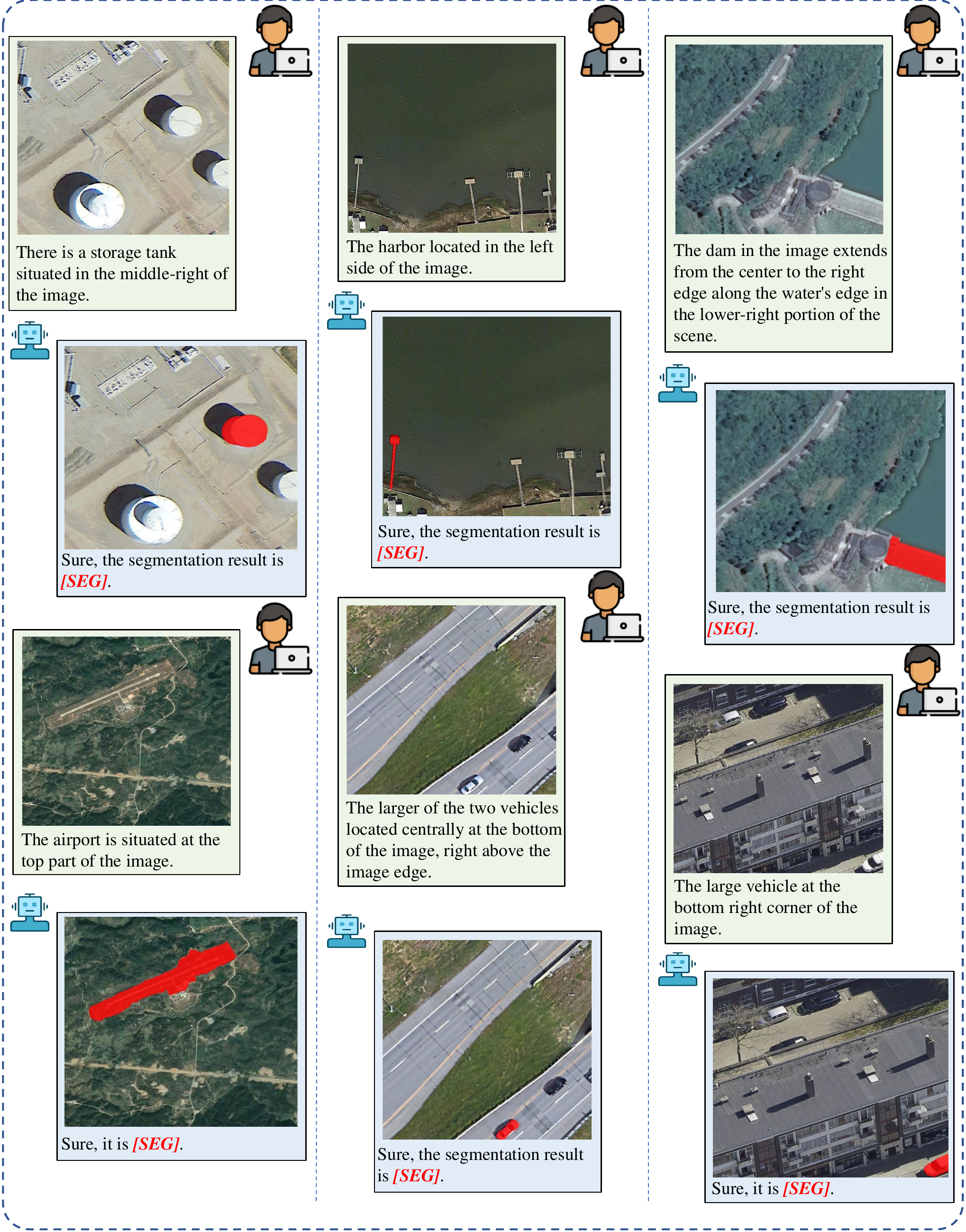}
    \caption{Visualization results of SegEarth-R2 on the RISBench test set.}
    \label{fig:qualitative_results_risbench}
\end{figure*}

Figure~\ref{fig:qualitative_results_refsegrs}, Figure~\ref{fig:qualitative_results_rrsis-d}, and Figure~\ref{fig:qualitative_results_risbench} show the qualitative results on the RefSegRS, RRSIS-D, and RISBench benchmarks, respectively.


\begin{figure*}[h]
    \centering
    \includegraphics[width=0.95\textwidth]
    {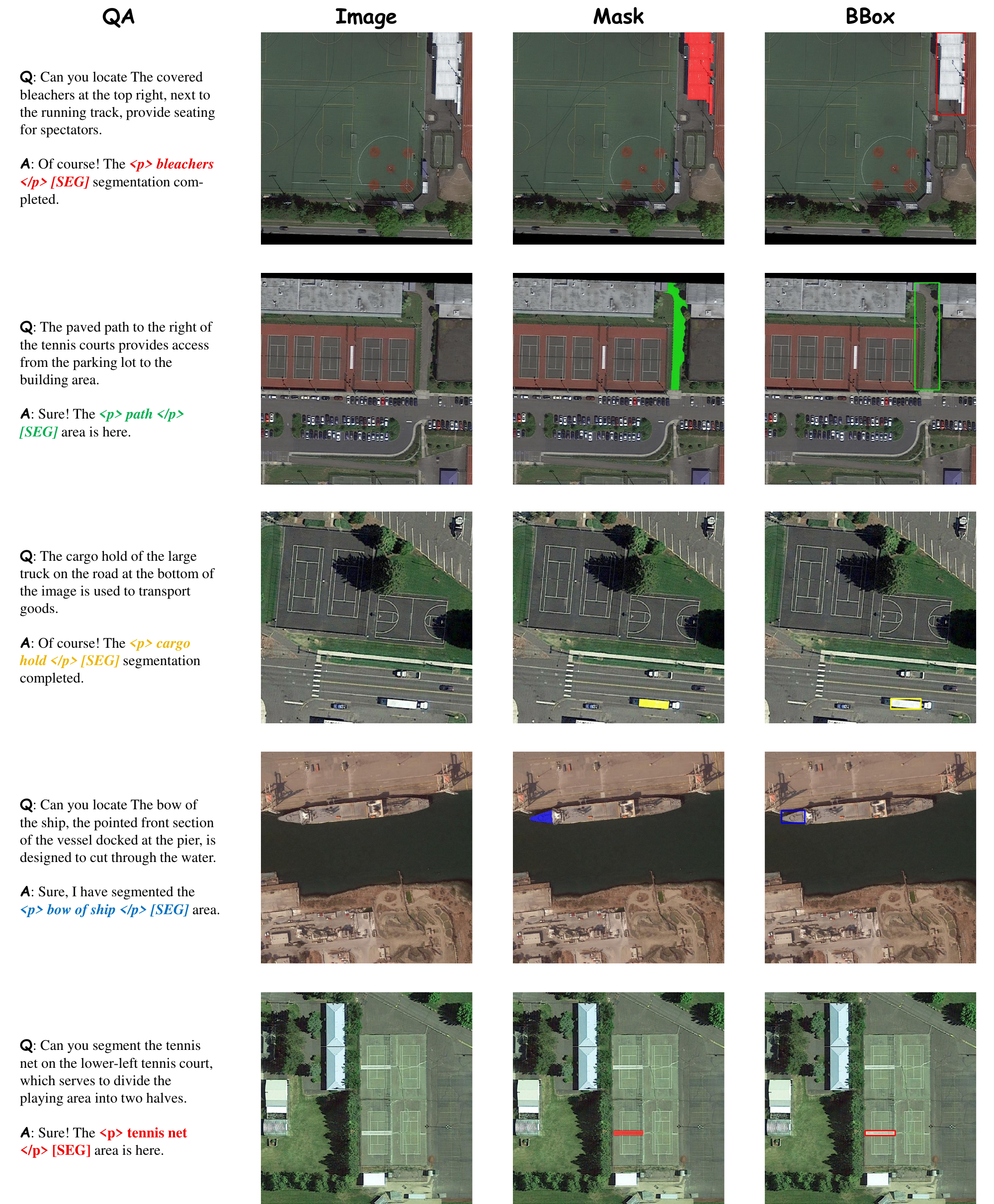}
    \caption{More examples of LaSeRS.}
    \label{fig:appendix_examples_3}
\end{figure*}


\begin{figure*}[h]
    \centering
    \includegraphics[width=0.95\textwidth]
    {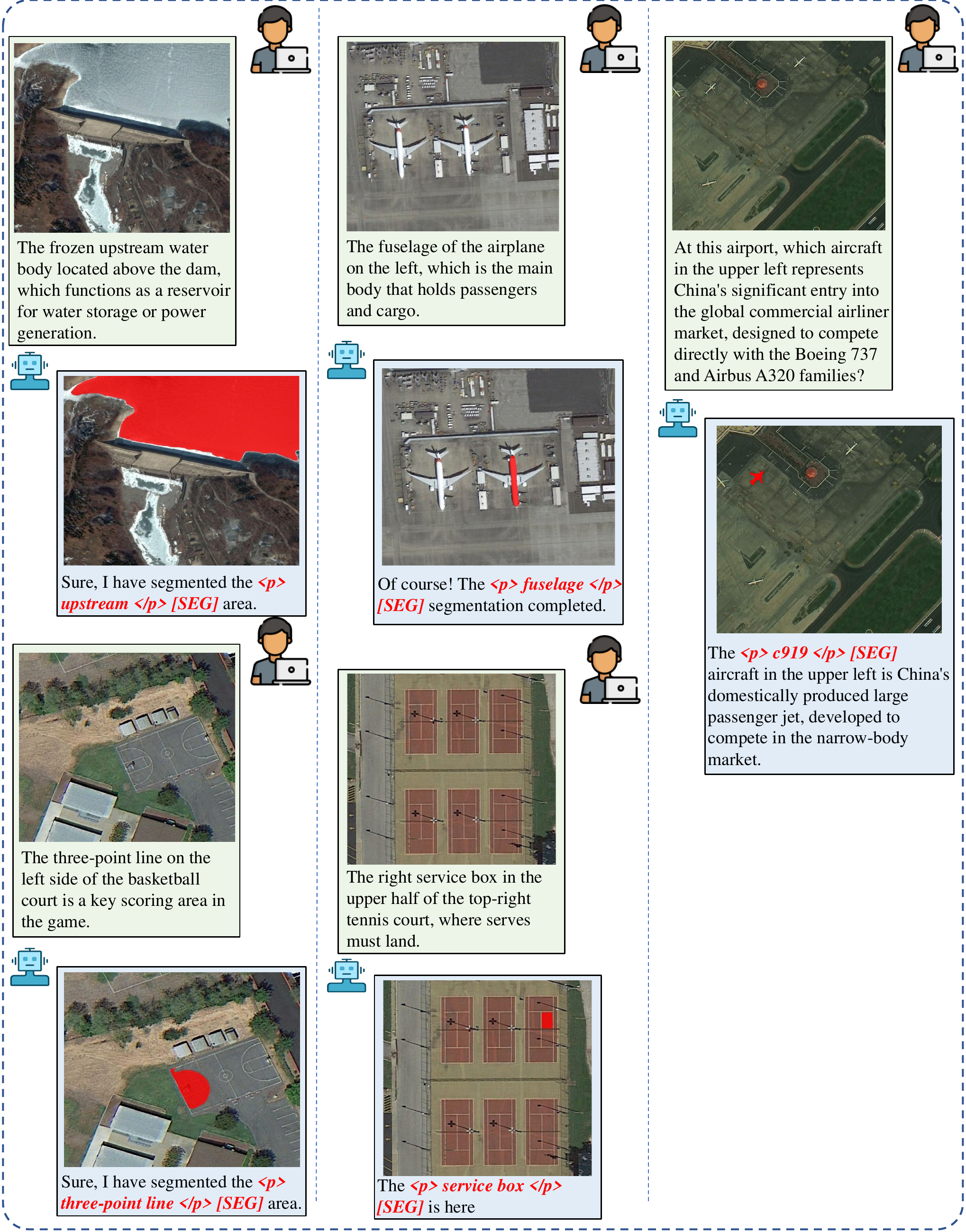}
    \caption{Visualization results of SegEarth-R2 on the LaSeRS dataset.}
    \label{fig:qualitative_results_2}
\end{figure*}

\subsection{EarthReason}
\label{sec:earthreason}
\begin{figure*}[h]
    \centering
    \includegraphics[width=0.95\textwidth]
    {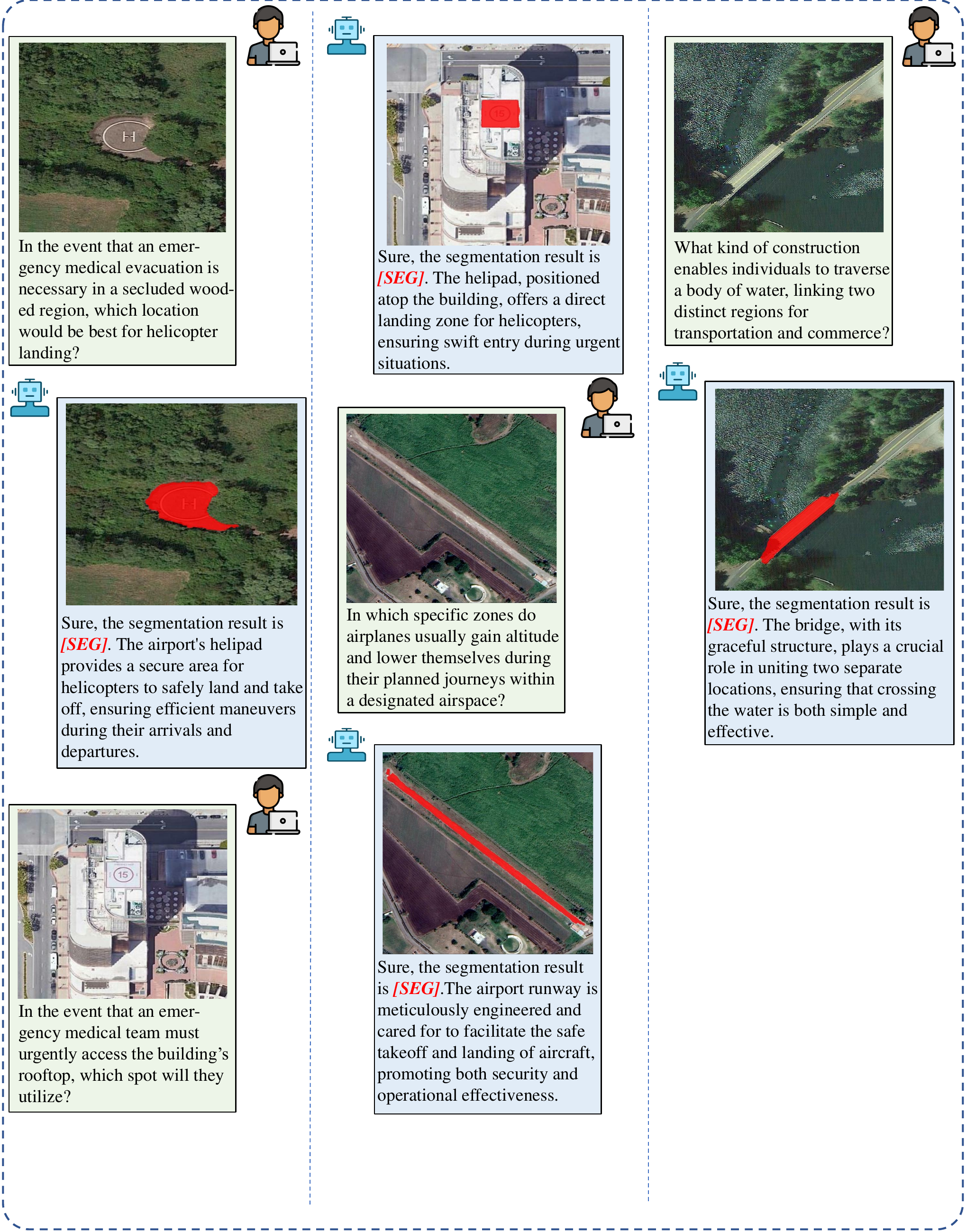}
    \caption{Visualization results of SegEarth-R2 on the EarthReason test set.}
    \label{fig:qualitative_results_earthreason}
\end{figure*}

Figure~\ref{fig:qualitative_results_earthreason} show the qualitative results on EarthReason benchmark.

\end{document}